\documentclass[runningheads]{llncs}
\usepackage{graphicx}

\usepackage{tikz}
\usepackage{comment}
\usepackage{amsmath,amssymb} 
\usepackage{color}
\usepackage{orcidlink}
\usepackage{booktabs}
\usepackage{multirow}
\usepackage{cite}

\usepackage[accsupp]{axessibility}  

\begin{document}
\pagestyle{headings}
\mainmatter
\def\ECCVSubNumber{6565}  

\title{DProST: Dynamic Projective Spatial Transformer Network for 6D Pose Estimation} 

\titlerunning{DProST: Dynamic Projective Spatial Transformer Network}
%
\author{Jaewoo Park\inst{1,2}\orcidlink{0000-0002-6816-4381} \and
Nam Ik Cho\inst{1,2,3}\orcidlink{0000-0001-5297-4649}}
\authorrunning{J. Park et al.}
%
\institute{Department of ECE \& INMC, Seoul National University, Seoul, Korea \and 
SNU-LG AI Research Center, Seoul, Korea \and
IPAI, Seoul National University, Seoul, Korea \\
\email{\{bjw0611,nicho\}@snu.ac.kr}}

\maketitle

\begin{abstract}
Predicting the object's 6D pose from a single RGB image is a fundamental computer vision task. Generally, the distance between transformed object vertices is employed as an objective function for pose estimation methods. However, projective geometry in the camera space is not considered in those methods and causes performance degradation. In this regard, we propose a new pose estimation system based on a projective grid instead of object vertices. Our pose estimation method, dynamic projective spatial transformer network (DProST), localizes the region of interest grid on the rays in camera space and transforms the grid to object space by estimated pose. The transformed grid is used as both a sampling grid and a new criterion of the estimated pose. Additionally, because DProST does not require object vertices, our method can be used in a mesh-less setting by replacing the mesh with a reconstructed feature. Experimental results show that mesh-less DProST outperforms the state-of-the-art mesh-based methods on the LINEMOD and LINEMOD-OCCLUSION dataset, and shows competitive performance on the YCBV dataset with mesh data. The source code is available at \url{https://github.com/parkjaewoo0611/DProST}.

\keywords{6D Pose Estimation; Spatial Transformer Network; 3D Reconstruction}
\end{abstract}

\section{Introduction}

Single image object pose estimation attempts to predict the transformation from the object space to the camera space based on an observed RGB image. Because the transformation can be expressed as rotation and translation, each having three degrees of freedom (DoF), it is also called the 6-DoF pose estimation. Finding the pose is commonly required in augmented reality (AR) \cite{marchand2015pose}, robot grasping problems \cite{cheng20216d,tremblay2018deep,zhu2014single,wang2019densefusion}, and autonomous driving \cite{chen2017multi,xu2018pointfusion}.

\begin{figure}[t]
\centering
\includegraphics[width=0.49\columnwidth]{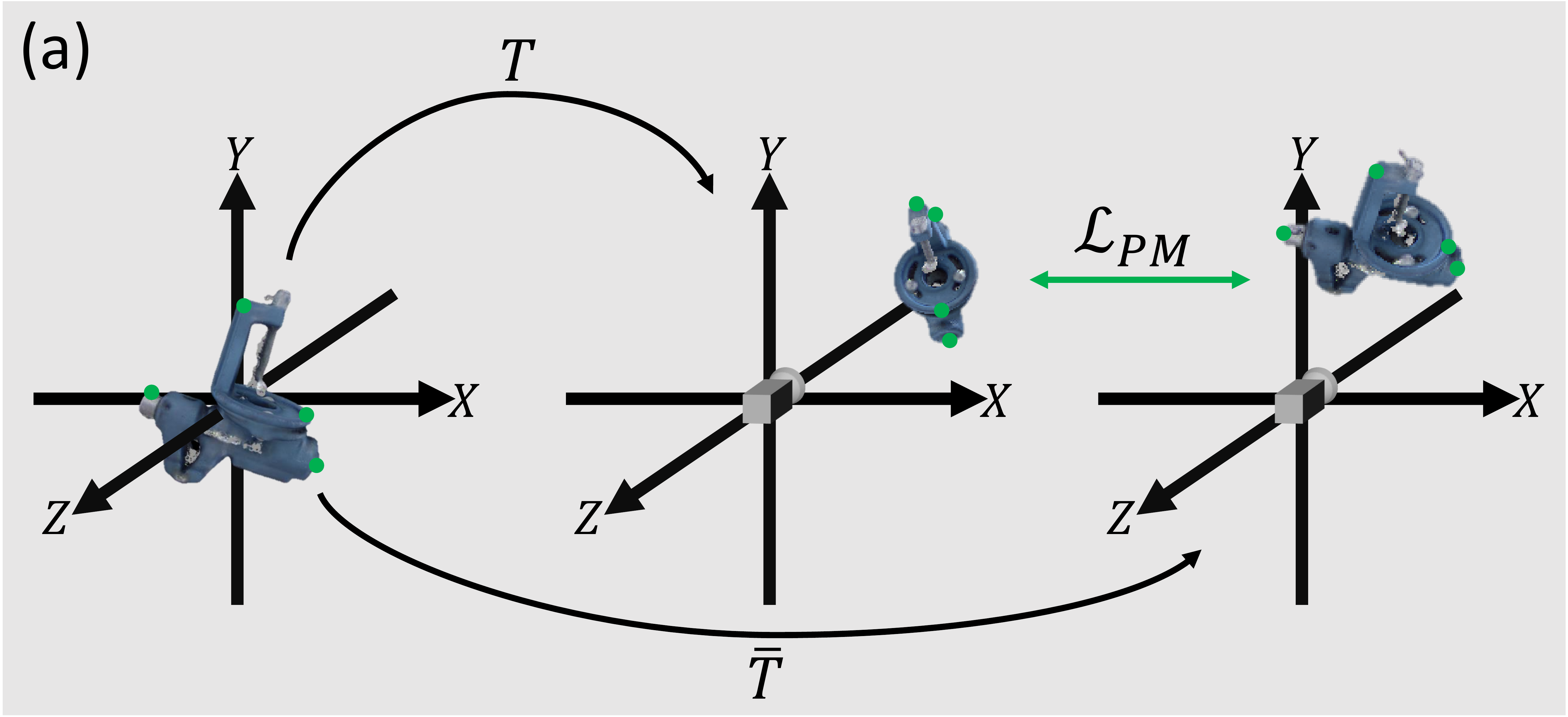}
\includegraphics[width=0.49\columnwidth]{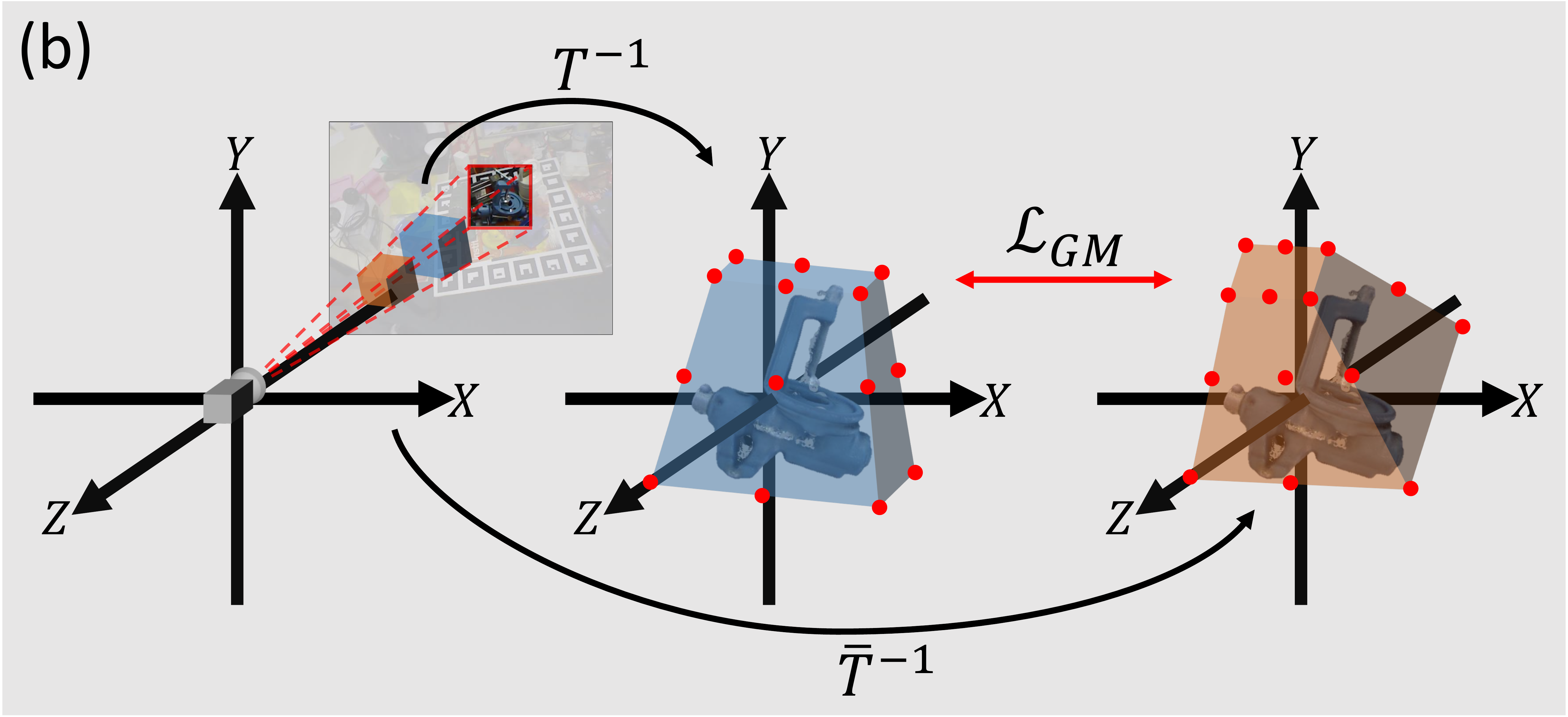}
\includegraphics[width=0.99\columnwidth]{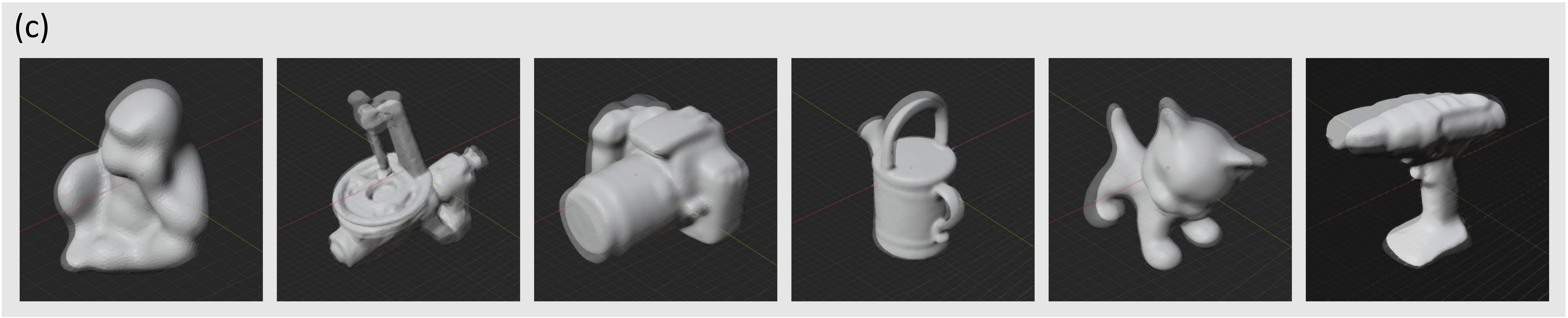}
\caption{\textbf{Key idea and perspective effect}: (a) Point matching loss is based on the distance between object vertices in camera space. The estimated pose is represented by $T$, while the ground-truth pose is denoted by $\bar{T}$. (b) Grid matching loss is based on the distance between cone-beam grid in object space. The region of interest grid is localized in camera space and transformed to object space by inverse transformation $T^{-1}$. We colorize the grid in blue and brown for each prediction and ground-truth. (c) The superimposed examples of each perspective and orthographic projection visualize the shape variance due to projective geometry.}
\label{fig:keyidea}
\end{figure}

Recently, deep learning-based methods have shown outstanding performance in complex computer vision problems. Therefore, researchers have proposed methods for applying deep learning to the object pose estimation problem with great performance \cite{oberweger2018making,peng2019pvnet,song2020hybridpose,rad2017bb8,hu2019segmentation,hu2020single,brachmann2016uncertainty,park2019pix2pose,wang2019normalized,li2019cdpn,wang2021gdr,di2021so,li2018deepim,labbe2020cosypose}. Some of these state-of-the-art methods use the point matching (PM) loss for pose estimation \cite{li2018deepim,labbe2020cosypose}. In particular, they used the pair-wise distance of the transformed object vertices by predicted pose and ground-truth pose in camera space.
The PM loss is a trivial approach in the camera space. However, although the PM-based methods assume the camera's focal length and principal point available, the perspective effect is ignored because they only consider the vertices on the rigid object as shown in Fig. \ref{fig:keyidea}(a). Consequently, the perspective effect is treated as a noisy shape variance as shown in Fig. \ref{fig:keyidea}(c) that degrades the pose estimation performance. On the other hand, image-space representation based methods \cite{chen2020category,peng2019pvnet,park2019pix2pose,wang2019normalized,li2019cdpn,park2020neural} may learn the perspective effect implicitly by the projected image. However, they also fail to address the projective geometry in their loss function and suffer from z-direction translation error because of 3D information loss.

Meanwhile, projective spatial transformer (ProST) network has been proposed in \cite{gao2020generalizing}, which first considered the projective geometry in spatial transformer network \cite{jaderberg2015spatial} to reflect the perspective effect. As the target of ProST is CT/radiograph registration, it dealt with only the limited camera pose, where the distance from the camera to the view frustum is fixed. However, since the region-of-interest (RoI) on camera space is vastly distributed in the object pose estimation, the ProST approach is not directly applicable because of memory and computation issues. In other words, to apply ProST on the object pose estimation, dynamically focusing the grid on the localized RoI is necessary.
  
In this regard, we propose a dynamic projective spatial transformer network (DProST) that estimates object pose based on the localized RoI cone-beam grid covering the object in object space. This grid leverages the 3D information while considering the projective geometry, as shown in Fig. \ref{fig:keyidea}(b). Additionally, we propose the grid distance (GD) loss and the grid matching (GM) loss to train the model based on grid correspondence. The grid-based approach has four major advantages. First, the shape of the cone-beam grid reflects the projective geometry. Second, because the grid has 3D coordinates, grid matching shows accurate z-axis translation estimation. Third, because the grid is uniformly distributed, it is relatively free from the object shape biases. Finally, since it does not use object vertices, it can be applied in mesh-less settings and shows excellent performance even with a simple space carving-based voxel feature instead of mesh. We confirm that our method shows state-of-the-art performance in LINEMOD \cite{hinterstoisser2012model}, LINEMOD-OCCLUSION \cite{brachmann2014learning}, and comparable performance in YCBV \cite{Xiang2018posecnn} datasets.

Our contributions are summarized as follows:
\begin{itemize}
\item We propose DProST based on a localized grid in the object space for pose estimation.
\item We propose GM loss and GD loss considering projective geometry.
\item We confirm that our method can be used in a mesh-less environment based on space-carving feature that was extracted from reference images and masks.
\item We confirm that our method shows state-of-the-art performance on LINEMOD, LINEMOD-OCCLUSION, and competitive performance on YCBV benchmarks.
\end{itemize}

\section{Related Work}
Recently proposed deep learning-based single image object pose estimation methods can be divided into three types. The first is to estimate the 3D intermediate representation and then find the matching pose using the perspective-n-point (PnP) algorithm \cite{lepetit2009epnp}. For example, \cite{rad2017bb8,hu2019segmentation} used the corners of a 3D bounding box, \cite{peng2019pvnet,song2020hybridpose} detected projected 3D keypoints of an object, and \cite{park2019pix2pose} used 2D-3D coordinates to train the network. 

The second type used gradient updates to minimize the difference between latent features or projected texture. For example, \cite{iwase2021repose,chen2020category} updated the pose by minimizing the difference between 2D reconstruction result and the observed image. Furthermore, as sophisticated novel view synthesis methods like \cite{RenderNet2018,sitzmann2019deepvoxels,mildenhall2020nerf} are proposed, pose estimation methods based on 3D view projection such as \cite{yen2021inerf,park2020latentfusion} are also suggested, which also use the gradient update over the view projection models. In particular, \cite{park2020latentfusion} shows both RGB-based and RGBD-based results, and considers mesh-less unseen object scenario. However, view projection-based methods require a lot of computation overhead to learn the 3D feature. Additionally, although an unseen and mesh-less scenario in \cite{park2020latentfusion} has great generalizability, it is not as accurate as the state-of-the-art methods based on mesh-based seen object. Considering the pros and cons of the above-referenced methods, to improve the generalizability without compromising the performance, we focus on mesh-less seen object scenario by replacing the mesh with the reference feature. Additionally, to reduce the computation overhead of 3D reconstruction, we use a simple space-carving method instead of a deep learning model. 

The last type directly estimates the pose from the network. For example, \cite{li2019cdpn,zakharov2019dpod,wang2021gdr,di2021so} proposed a learning method to use both 2D-3D coordinate representation and direct pose regression. \cite{li2018deepim} used the rendered object image and the observed object image as inputs to iteratively refine the pose. Also, in this method, the disentangled representation of rotation and translation greatly improved the performance. Based on an iterative fashion and disentangled representation similar to \cite{li2018deepim}, the single view method in \cite{labbe2020cosypose} showed the most superior performance in the object pose estimation challenge \cite{hodavn2020bop} using a large synthetic image dataset. However, because these methods are based on point matching loss, they fail to address the projective geometry in objective function. We combine the iterative pose refining configuration with ProST \cite{gao2020generalizing} to consider both projective geometry and the performance.

\section{Method}
\subsection{Framework Overview}

\begin{figure}[t]
\centering
\includegraphics[width=1.0\columnwidth]{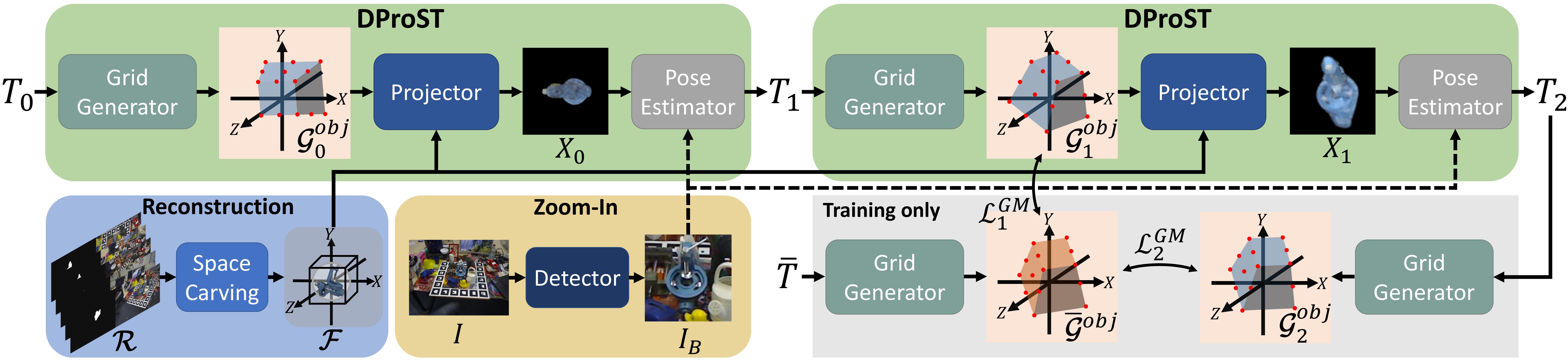}
\caption{\textbf{Overview of DProST}: Space carving extracts a reference feature from images and masks. Given the reference feature and detected object, the object's pose is iteratively refined by DProST. Grid generator converts the pose to the object space grid visualized in red dots. The projector outputs estimated appearance based on the grid and the reference feature. Then, the pose estimator refines the pose based on appearance. Grid is also used as a comparison target for GM loss during the training phase.}
\label{fig:overview}
\end{figure}

The overall process of DProST is shown in Fig. \ref{fig:overview}. We follow the zoom-in setting used in \cite{li2019cdpn,labbe2020cosypose,li2018deepim}. Hence, the off-the-shelf detector is first used to find the bounding box area of the object $I_B$ from the overall image $I$, where the bounding box is $B=(x, y, w, h)$. 

Then, the space-carving-based reconstruction stage generates a reference feature $\mathcal{F}$ in the object space's unit sphere based on the reference set $\mathcal{R}=\left\{\left(I^{\mathcal{R}}_{k}, M^{\mathcal{R}}_{k}, \bar{T}^{\mathcal{R}}_{k}, K^{\mathcal{R}}_{k} \right) | k \in \left[1, N^{\mathcal{R}}\right]\right\}$ consists of $N^{\mathcal{R}}$ tuples of images $I^{\mathcal{R}}_{k}$ and masks $M^{\mathcal{R}}_{k}$ with known poses $\bar{T}^{\mathcal{R}}_{k}$ and intrinsic matrices $K^{\mathcal{R}}_{k}$ sampled from training set, which can be written as
\begin{align}
\mathcal{F} = f_{carv}(\mathcal{R}).
\end{align} 
We will discuss the details about the reconstruction stage in Section 3.3.

Subsequently, we iteratively refine the initial pose $T_0$ based on the $\mathcal{F}$ and $I_B$.
Each pose refinement iteration $i$ consists of three submodules. First, based on a bounding box $B$, object pose $T_i=\{R_i, t_i\}$, and intrinsic camera parameter $K$, grid generator extracts a localized RoI grid as 
\begin{align}
\mathcal{G}_i^{obj} = f_{grid}(B, T_i, K)
\end{align} 
on the rays from the camera to the image plane in object space. Grid generator is composed of four sub-steps, grid forming, grid cropping, grid pushing, and grid transformation, and we will discuss the details of each step in Section 3.2. 

The projector renders object appearance $X_i$ of each iteration by sampling features from $\mathcal{F}$ based on $\mathcal{G}_i^{obj}$ that can be written as 
\begin{align}
X_i = f_{proj}(\mathcal{G}_i^{obj}, \mathcal{F}).
\end{align} 
The projection stage plays a similar role to mesh rendering in \cite{li2018deepim,labbe2020cosypose}. It can also be replaced with mesh rendering methods when the reference pose is not diverse enough to carve the object shape in the reconstruction stage. 

Finally, to refine the pose, the pose estimator network of the $i$-th iteration with parameter $\theta_i$ predicts the relative pose $\Delta T_{i}$ between $X_i$ and $I_B$ as 
\begin{align}
\Delta T_{i} = f_{\theta_i}(X_i, I_B).
\end{align} 
The details of the projection stage and pose estimator are discussed in Section 3.4. Finally, we will address the overall objective function in Section 3.5.  

\subsection{Object Space Grid Generator}

\begin{figure}[t]
\centering
\includegraphics[width=1.0\textwidth]{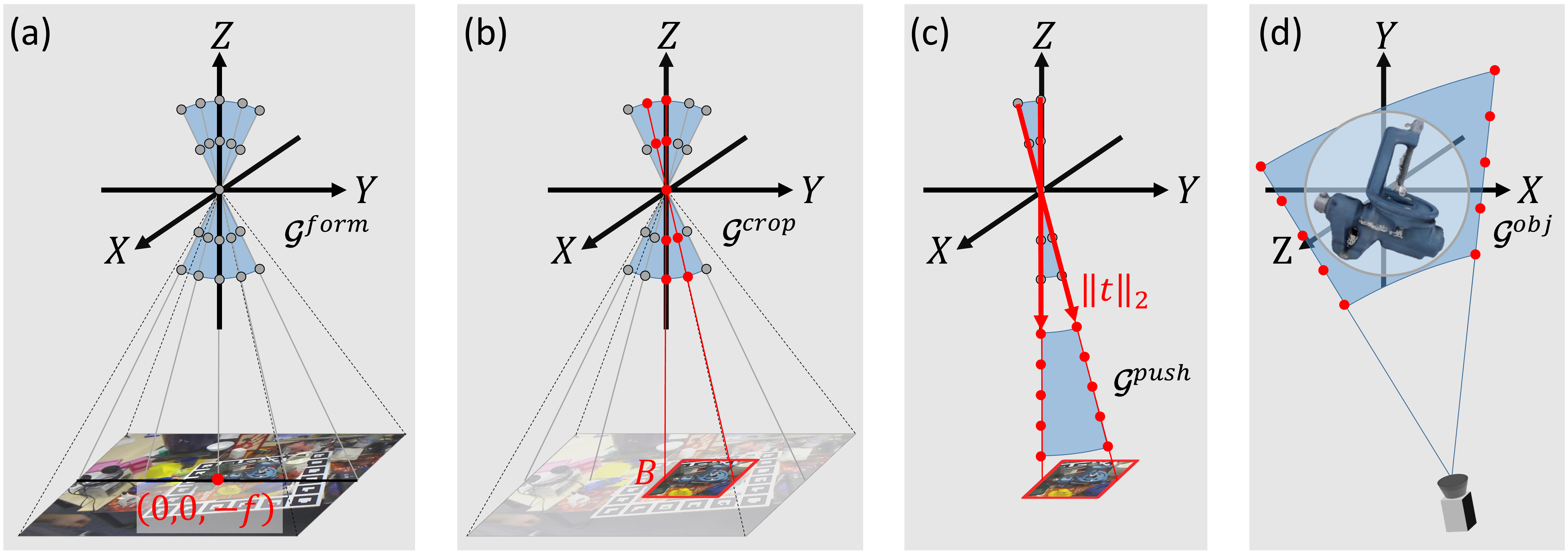}
\caption{\textbf{Substeps in grid generator}: (a) The grid inside a unit sphere is generated in camera space on the ray to the 3D image plane. (b) The RoI grid is extracted based on the bounding box. (c) The grid is pushed by $\|t\|_{2}$ along the ray to cover the object in camera space. (d) The grid is transformed by $T^{-1}$ from camera space to object space.}
\label{fig:grid_generator}
\end{figure}

Grid generator in DProST converts the given intrinsic matrix ($K$), object pose ($T$), and bounding box ($B$) to the RoI grid in object space ($\mathcal{G}^{obj}$). Two key ideas are used for RoI grid localization. First, as the object's projection is included in the bounding box, the object in camera space is in the ray's direction, which passes through the bounding box area. Second, the distance to the object is the size of the translation vector in object pose.

\subsubsection{Grid Forming:} We first generate the grid on the rays from the camera (origin) to the image plane pixel's 3D location $\mathcal{I}\in \mathbb{R}^{H \times W \times 3}$ in the camera space and match the principal point to $(0, 0, -f)$ as shown in Fig. \ref{fig:grid_generator}(a). The initialized grid with the given the focal length $f$ and principal point $\textbf{p}=(p_x, p_y)$ can be written as follows:
\begin{gather}
\mathcal{I}\left(l, m\right) = (l-p_x, m-p_y, -f), \\ 
\mathcal{G}^{form}\left(l,m,n\right) = \frac{\mathcal{I}\left(l, m\right)}{\|\mathcal{I}\left(l, m\right)\|_{2}} \left(\frac{2n}{N_z}-1\right),
\end{gather}
where $\left(l, m\right)$ is the $\left(x, y\right)$ index of a pixel in $\mathcal{I}$, the $\mathcal{G}^{form} \in \mathbb{R}^{H \times W \times N_z \times 3}$ is formed grid, $N_z$ is the number of points in the grid on each ray, and $n \in [0, N_z - 1]$ is the index of the points on each ray, respectively.

\subsubsection{Grid Cropping:} Bounding box is used as a direction indicator from camera to object in camera space. As shown in Fig. \ref{fig:grid_generator}(b), we used the RoI align \cite{he2017mask} method to extract the grid on the rays that are projected into the bounding box $B=(x,y,w,h)$ to focus the grid near the object, which can be written as   
\begin{align}
\mathcal{G}^{crop} = RoIAlign\left(\mathcal{G}^{form}, B \right),
\label{equ:grid_crop} 
\end{align}
where $\mathcal{G}^{crop} \in \mathbb{R}^{h \times w \times N_z \times 3}$ is the extracted RoI grid.

\subsubsection{Grid Pushing:} We then push the grid on the rays as much as the distance to object $\|t\|_{2}$ towards the $\mathcal{I}$ to cover the object in camera space, as shown in Fig. \ref{fig:grid_generator}(c), which can be written as 
\begin{align}
\mathcal{G}^{push} = \big\{g - \|t\|_{2} \frac{g}{\|g\|_2} sign\left(\left(g\right)_z\right) | g \in \mathcal{G}^{crop}\left(l, m, n\right)\},
\label{equ:grid_push} 
\end{align}
where $\mathcal{G}^{push} \in \mathbb{R}^{h \times w \times N_z \times 3}$ is the pushed grid in camera space, and the $sign$ function is used to invert the pushing direction of grid in $+Z$ to $-Z$ direction.

\subsubsection{Grid Transformation:} Finally, to cover the $\mathcal{F}$ in object space with the grid, as shown in Fig. \ref{fig:grid_generator}(d), we transform the pushed grid from camera space to object space with inverse of pose transformation as 
\begin{align}
\mathcal{G}^{obj} = T^{-1}\left( \mathcal{G}^{push} \right),
\label{equ:grid_transform} 
\end{align}
where $\mathcal{G}^{obj} \in \mathbb{R}^{h \times w \times N_z \times 3}$ is the RoI grid that tightly wraps around the $\mathcal{F}$. Note that unlike point matching-based and render-and-compare-based methods \cite{labbe2020cosypose,li2018deepim}, which require transformation twice for each loss computation and rendering,  $\mathcal{G}^{obj}$ can be used for both grid sampling in projector and loss function in our method.

\subsection{3D Feature Reconstruction}

\begin{figure}[t]
\centering
\includegraphics[width=1.0\columnwidth]{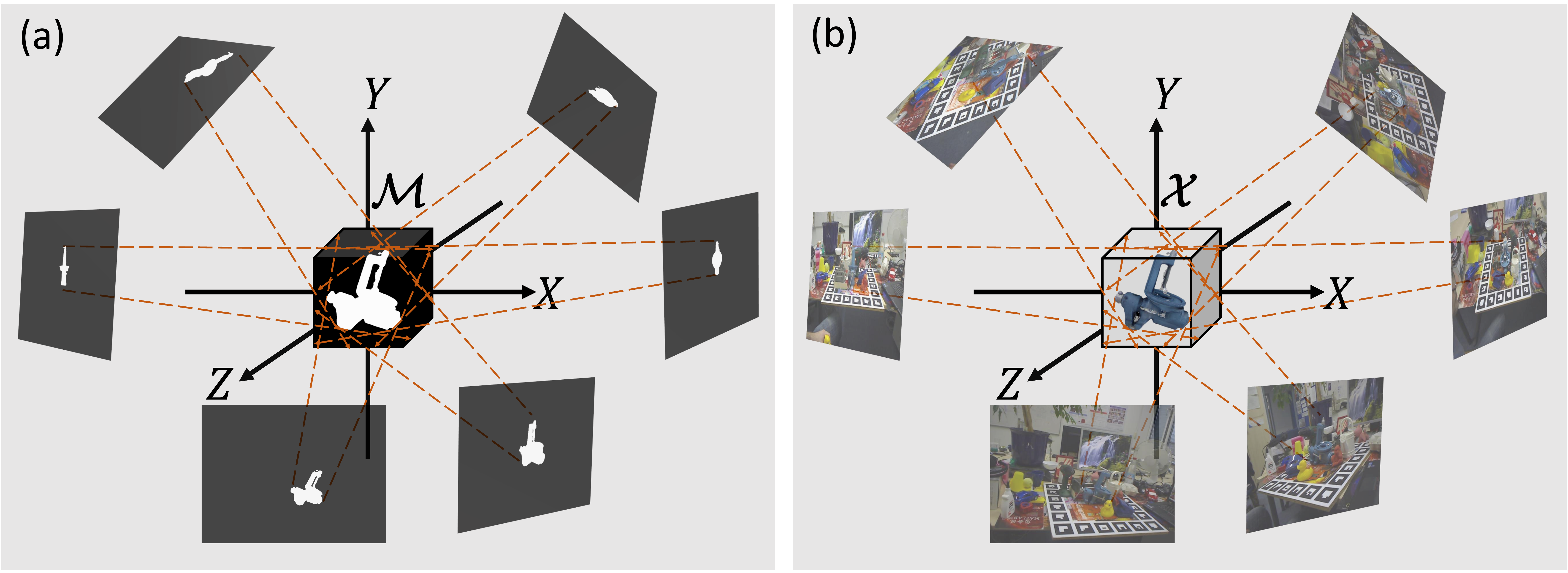} 
\caption{\textbf{3D Feature reconstruction}: We visualize the object shaping and coloring process of generating reference feature. (a) Only the voxels projected into each reference object mask remain positive, while others are cut off. (b) The values in reference images are averaged to extract the RGB value of reference voxels.}
\label{fig:space_carving}
\end{figure}

In this section, we present a non-learning-required and a simple  space-carving-based reference feature generation method. To generate the reference feature $\mathcal{F}$ in the unit sphere of the object space, we use pose-known reference images $\{X^{k} | k \in [1, N^{\mathcal{R}}]\}$ and corresponding masks $\{M^{k} | k \in [1, N^{\mathcal{R}}]\}$ sampled from the training set. 
The reference quality is improved when the various views and fine details are included. Hence, we set the first reference image as the sample with the largest object mask in the training set. Then, based on the geodesic distance of the rotation matrix, we use the farthest point sampling (FPS) algorithm in \cite{peng2019pvnet} to select the other reference images. 
We generate a 3D canvas as a normalized voxel grid $\mathcal{A} \in \mathbb{R}^{S \times S \times S \times 3}$, which is divided uniformly into $S^3$ voxels, and each contains the coordinates of itself. Then, for each reference mask and image, $\mathcal{A}$ is projected with the given $K$ and the pose of the $k$-th reference $\bar{T}^{k}$ to find the region where the canvas grid is projected. Hence, we assign the projected canvas location in $A^{k}$ as 
\begin{align}
A^{k} = \pi \left( \bar{T}^{k}\left( \mathcal{A} \right), K \right),
\label{equ:ref_projection}
\end{align}
where $\pi(p, K)$ represents the projection of the point $p$ with the intrinsic matrix $K$ in the pinhole camera model.
Then, using $A^{k}$ as a grid, we apply the grid sampling method proposed in \cite{jaderberg2015spatial} to generate a 3D RGB feature $\mathcal{X}^{k} \in \mathbb{R}^{S  \times S \times S \times 3} $ from $X^{k}$ and a 3D mask feature $\mathcal{M}^{k} \in \mathbb{R}^{S  \times S \times S \times 1} $ from $M^{k}$ as follows:
\begin{align}
\mathcal{X}^{k} = X^{k}\left( \left(a^{k}\right)_{x}, \left(a^{k}\right)_{y} \right), \\
\mathcal{M}^{k} = M^{k}\left( \left(a^{k}\right)_{x}, \left(a^{k}\right)_{y} \right),
\label{equ:ref_grid_sample} 
\end{align}
where $\left(a^{k}\right)_{x}$ and $\left(a^{k}\right)_{y}$ are $\left(x, y\right)$ coordinates in $A^{k}$. Note that we use nearest-neighbor interpolation to extract values from non-integer locations. Then, we average the $\mathcal{X}^{k \in [1, N^{\mathcal{R}}]}$'s to obtain the integrated 3D RGB feature $\mathcal{X}$, and multiply $\mathcal{M}^{k \in [1, N^{\mathcal{R}}]}$'s to get the shape-carved 3D mask feature $\mathcal{M}$. Finally, the 3D features of the object are calculated as 
\begin{align}
\mathcal{F} = \mathcal{M} \odot \mathcal{X},
\label{equ:3D_feature} 
\end{align}
where $\mathcal{F} \in \mathbb{R}^{S \times S \times S \times 3} $ is an RGB 3D feature with a carved shape with the object information in voxel fashion, and $\odot$ is an element-wise multiplication. The concept of carving process is visualized in Fig. \ref{fig:space_carving}.

\subsection{Projector and Pose Estimator}
With the $\mathcal{G}^{obj}$ obtained from grid generator and the reference feature $\mathcal{F}$, we apply the grid sampling \cite{gao2020generalizing}, which can be written as 
\begin{align}
\mathcal{F}^{interp} = interp(\mathcal{F}, \mathcal{G}^{obj}),
\end{align}
where $\mathcal{F}^{interp} \in \mathbb{R}^{h \times w \times N_z \times 3}$ is the sampled feature of the object and $interp$ is the trilinear interpolation-based grid sampling function. 

Then, to compare the estimated pose and $I_B$, we project $\mathcal{F}^{interp}$ into a 2D feature $F \in \mathbb{R}^{h \times w \times 3}$ by choosing the closest valid point to the camera on each ray as 
\begin{align}
F = \mathcal{F}^{interp} \left(l, m, \min_{n} \{ \delta\left(l, m \right)\}  \right),
\label{equ:Z-buffering} 
\end{align}
where $\delta\left(l, m \right) =  \{ n | \mathcal{F}^{interp} \left( l, m, n \right) \neq 0 \}$. 

For each iteration $i$, we use a ResNet34 \cite{he2016deep} based pose estimator to refine the pose $T_{i-1}$ to $T_i$ to match the projected reference $F_{i-1}$ to the detected object image $I_B$. The last layer of ResNet34 is replaced with a fully connected layer that outputs nine values, each of which is image space translation, scale factor, and two axis-based relative rotation representation as in \cite{labbe2020cosypose}. See the supplementary materials for more details about the pose estimator's output format.

\subsection{Objective Functions}
Instead of PM loss based on the mesh vertices of the object, we propose GM loss for mesh-less training. For each iteration $i$, we use the Euclidean distance of the object space grid as the GM loss, which can be written as
\begin{align}
\mathcal{L}^{GM}_i = \frac{1}{|\mathcal{G}_i^{obj}|} \sum \left\|\mathcal{\bar{G}}^{obj} - \mathcal{G}^{obj}_{i}\right\|_{2},
\label{equ:GM} 
\end{align}
where $|\mathcal{G}^{obj}_i|$ is the number of points in the grid, $\mathcal{\bar{G}}^{obj}$ is a grid from the ground-truth pose $\bar{T}$, and $\mathcal{G}^{obj}_{i}$ is the predicted object space grid. Note that, as the DProST module is based on fully differentiable operations, the $i$-th GM loss is used as an objective function for the $i$-th pose estimator, as shown in Fig. \ref{fig:overview}.

Additionally, to help the pose estimator determine the distance from the camera to the object in the grid pushing step, we use an auxiliary loss called GD loss as
\begin{align}
\mathcal{L}^{GD}_{i} = \big\|\|\bar{t}\|_{2} - \|t_{i}\|_{2}\big\|_{1},
\label{equ:GD} 
\end{align}
which is the absolute difference between the predicted distance and ground-truth distance in the grid pushing step.

Finally, the overall loss $\mathcal{L}_i$ for training our network on each iteration $i$ is the combination of the above two losses as
\begin{align}
\mathcal{L}_i= \mathcal{L}^{GM}_{i} + \lambda^{GD}\mathcal{L}^{GD}_{i},
\label{equ:total_loss} 
\end{align}
where $\lambda^{GD}$ is the balance factor. The $\mathcal{L}_i$ is applied to outputs of each iterations' pose estimator.

\section{Experiments}
\subsection{Datasets and Evaluation Metrics} The performance is tested on LINEMOD (LM) \cite{hinterstoisser2012model}, LINEMOD-OCCLUSION (LMO) \cite{brachmann2014learning}, and YCB-video (YCBV) dataset \cite{Xiang2018posecnn}. The LM consists of 13 objects with approximately 1.2K images per object. We follow the settings described in \cite{brachmann2016uncertainty}, which uses 15\% of the data for training and the rest for testing. The LMO dataset is a subset of the LM dataset consisting of eight objects in more cluttered scenes. The YCBV dataset consists of 21 objects with between 10k and 20k real images per object. The YCBV dataset is very hard compared to the LM dataset because of the hard occluded samples. Also, the views in YCBV dataset are not as various as the LM dataset. We used the official pbr dataset from the BOP challenge \cite{hodavn2020bop} in training along with the real image training set. For the LM dataset, we also report the performance trained on the synthetic training dataset introduced from \cite{li2018deepim}.

The $ADD(\textbf{-}S)$ score \cite{hinterstoisser2012model} is used to measure the performance, which is the most widely employed metric in the object pose estimation \cite{Xiang2018posecnn,li2018deepim,song2020hybridpose,wang2021gdr,song2020hybridpose,iwase2021repose}. In particular, we use the $ADD$ score for non-symmetric objects, which computes the pair-wise distance of vertices transformed by the predicted pose and ground-truth pose. And we use the $ADD\textbf{-}S$ score for symmetric objects, which measures the nearest distance of transformed vertices by predicted pose and ground-truth pose.
The predicted pose is considered correct if the mean distance is less than the threshold ratio $thr$ of the object diameter in the $ADD(\textbf{-}S)^{thr}$ metric. The area-under-curve of the $ADD(\textbf{-}S)$ metric \cite{Xiang2018posecnn} is additionally used for the YCBV dataset.

We also compare the projection error $Proj2D$ \cite{li2018deepim}, which measures the pixel-level discrepancy of the projected vertices in the image space.

\subsection{Experiment Setup}
\subsubsection{Implementation Details.} Our pipeline is implemented based on the Pytorch \cite{paszke2019pytorch} and Pytorch3d \cite{ravi2020accelerating} framework. We used the ADAM optimizer \cite{kingma2014adam} with a learning rate of 0.0001. Because the number of real training images is much smaller than the number of synthetic training images, we construct half of each mini-batch from real images and the other half from synthetic images to focus on the real dataset. The pose estimator is trained for 3,000 epochs based on the real dataset, and we divide the learning rate by ten on the 2,000th epoch for the LM and LMO dataset. Similarly, we train on the YCBV dataset for 300 epochs and divide the learning rate on the 200th epoch.

The pretrained ResNet34 \cite{he2016deep} on ImageNet \cite{deng2009imagenet} is used as the backbone of the pose estimator. To prevent the object from being out of sight in the initial pose, we set the initial translation vector to fit the projection of the 3D reference feature into the bounding box with the identity matrix as the initial pose's rotation as implemented in \cite{labbe2020cosypose}. For more details about the initial pose, check the supplementary materials. Additionally, we use distinct weights for the pose estimator in each iteration to teach the models to work in a cascade fashion \cite{cai2018cascade}, that shows better performance than that of iterative models. We use $\lambda^{GD}=1$ in Equation (\ref{equ:total_loss}). 
 
In the reconstruction stage, we use eight reference images $(N^{\mathcal{R}}=8)$ to generate reference features for each object and the number of voxels along the axis is 128 $(S=128)$ because we used a $128 \times 128$ image size for the pose estimator. In the grid generator, 64-grid points $(N_{z}=64)$ per ray are used and confirmed that the quality of the projection is saturated when $N_z$ is greater than 64. Details about hyperparameter experiments are given in the ablation study.

\subsubsection{Object Zoom-In.} We zoom in \cite{li2018deepim,labbe2020cosypose,li2019cdpn,wang2021gdr} the detected object to a fixed size $128 \times 128$, while keeping the aspect ratio of the image. In particular, we follow the dynamic zoom-in setting suggested in \cite{li2019cdpn}, where noisy ground-truth bounding boxes are used in the training phase and off-the-shelf detectors \cite{ren2015faster,redmon2018yolov3} are used in the test phase.

\begin{table}[t]
\caption{\textbf{Results on the LM dataset.} A comparison of other baseline methods and our method in terms of the $ADD(\textbf{-}S)^{0.1d}$ score. Objects annotated with $(*)$ indicate symmetric pose ambiguity. A.O represents a pose estimator trained on all objects.}
{\resizebox{1.0\columnwidth}{!}
{\begin{tabular}{c|cccccc|cccc}
\hline
\toprule
\multirow{2}{*}{Method}                                & \begin{tabular}[c]{@{}c@{}}PoseCNN\\ \cite{Xiang2018posecnn}\end{tabular} & \begin{tabular}[c]{@{}c@{}}DeepIM\\ \cite{li2018deepim}\end{tabular} & \begin{tabular}[c]{@{}c@{}}HybridPose\\ \cite{song2020hybridpose}\end{tabular} & \begin{tabular}[c]{@{}c@{}}GDR-Net\\ \cite{wang2021gdr}\end{tabular}  & \begin{tabular}[c]{@{}c@{}}SO-Pose\\ \cite{di2021so}\end{tabular} & \begin{tabular}[c]{@{}c@{}}RePOSE\\ \cite{iwase2021repose}\end{tabular} & DProST                                              & DProST                                              & DProST                                              & DProST                                              \\ \cline{2-11}
                                                       & \multicolumn{6}{c|}{w/ Mesh}                                                           & \multicolumn{4}{c}{w/o Mesh}                                                     \\ \hline
A.O                                                    & \checkmark                                                               & \checkmark                                                           &                                                                                & \checkmark                                                                       & \checkmark                                                          &                                                                     &                                                    & \checkmark                                                   &                                                     & \checkmark                                                   \\ \hline
\begin{tabular}[c]{@{}c@{}}Training\\ Set\end{tabular} & \begin{tabular}[c]{@{}c@{}}real\\ +syn\end{tabular}                       & \begin{tabular}[c]{@{}c@{}}real\\ +syn\end{tabular}                  & \begin{tabular}[c]{@{}c@{}}real\\ +syn\end{tabular}                            & \begin{tabular}[c]{@{}c@{}}real\\ +syn\end{tabular}                     & \begin{tabular}[c]{@{}c@{}}real\\ +syn\end{tabular}               & \begin{tabular}[c]{@{}c@{}}real\\ +syn\end{tabular}                  & \begin{tabular}[c]{@{}c@{}}real\\ +syn\end{tabular} & \begin{tabular}[c]{@{}c@{}}real\\ +syn\end{tabular} & \begin{tabular}[c]{@{}c@{}}real\\ +pbr\end{tabular} & \begin{tabular}[c]{@{}c@{}}real\\ +pbr\end{tabular} \\ \hline
ape                                                     & -                                                                         & 77.0                                                                 & 77.6                                                                           & -                                                                    & -                                                                 & 79.5                                                                    & 91.4                                                & 91.5                                                & 91.1                                                & \textbf{91.6}                                                \\
benchwise                                               & -                                                                         & 97.5                                                                 & 99.6                                                                           & -                                                                   & -                                                                 & \textbf{100.0}                                                                    & \textbf{100.0}                                               & 99.9                                                & \textbf{100.0}                                               & 99.7                                                \\
cam                                                     & -                                                                         & 93.5                                                                 & 95.9                                                                           & -                                                                    & -                                                                 & \textbf{99.2}                                                                    & 98.8                                                & 98.4                                                & 99.1                                                & 98.2                                                \\
can                                                     & -                                                                         & 96.5                                                                 & 93.6                                                                           & -                                                                    & -                                                                 & 99.8                                                                    & 99.5                                                & 99.5                                                & \textbf{99.9}                                               & 99.6                                                \\
cat                                                     & -                                                                         & 82.1                                                                 & 93.5                                                                           & -                                                                    & -                                                                 & 97.9                                                                    & 98.0                                                & 98.1                                                & 97.9                                                & \textbf{98.2}                                                \\
driller                                                 & -                                                                         & 95.0                                                                 & 97.2                                                                           & -                                                                    & -                                                                 & 99.0                                                                    & \textbf{99.5}                                                & 97.0                                                & 99.2                                                & 98.4                                                \\
duck                                                    & -                                                                         & 77.7                                                                 & 87.0                                                                           & -                                                                    & -                                                                 & 80.3                                                                    & 88.5                                                & \textbf{91.0}                                                & 89.2                                                & 90.8                                                \\
eggbox*                                                 & -                                                                         & 97.1                                                                 & 99.6                                                                           & -                                                                   & -                                                                 & \textbf{100.0}                                                                    & 99.9                                                & 99.8                                                & \textbf{100.0}                                               & 99.8                                                \\
glue*                                                   & -                                                                         & 99.4                                                                 & 98.7                                                                           & -                                                                    & -                                                                 & 98.3                                                                    & 99.8                                                & \textbf{100.0}                                               & \textbf{100.0}                                               & 99.9                                                \\
holepuncher                                             & -                                                                         & 52.8                                                                 & 92.5                                                                           & -                                                                    & -                                                                 & 96.9                                                                    & 97.0                                                & 97.0                                                & 96.4                                                & \textbf{97.6}                                                \\
iron                                                    & -                                                                         & 98.3                                                                 & 98.1                                                                           & -                                                                   & -                                                                 & \textbf{100.0}                                                                    & 99.5                                                & 98.8                                                & \textbf{100.0}                                                & 99.1                                                \\
lamp                                                    & -                                                                         & 97.5                                                                 & 96.9                                                                           & -                                                                    & -                                                                 & 99.8                                                                    & 99.6                                                & 99.7                                                & \textbf{100.0}                                               & 99.8                                                \\
phone                                                   & -                                                                         & 87.7                                                                 & 98.3                                                                           & -                                                                    & -                                                                 & \textbf{98.9}                                                                    & 95.8                                                & 98.2                                                & 97.4                                                & 97.7                                                \\ \hline
Average                                                 & 62.7                                                                      & 88.6                                                                 & 94.5                                                                           & 93.7                                                                    & 96.0                                                              & 96.1                                                                 & 97.5                                                & 97.6                                                & \textbf{97.7}                                                & \textbf{97.7}                                                \\ \bottomrule
\end{tabular}}}
\label{table:LINEMOD}
\end{table}

\subsection{Comparison with the State-of-the-Art}
Table \ref{table:LINEMOD} presents the results on the LM dataset. As shown in the table, our method shows state-of-the-art accuracy on the LM dataset even without mesh data. Compared to a point matching-based method \cite{li2018deepim}, which uses mesh vertices, our method shows more accurate results for almost every object. 

\begin{figure}[t]
\centering
\includegraphics[width=1.0\textwidth]{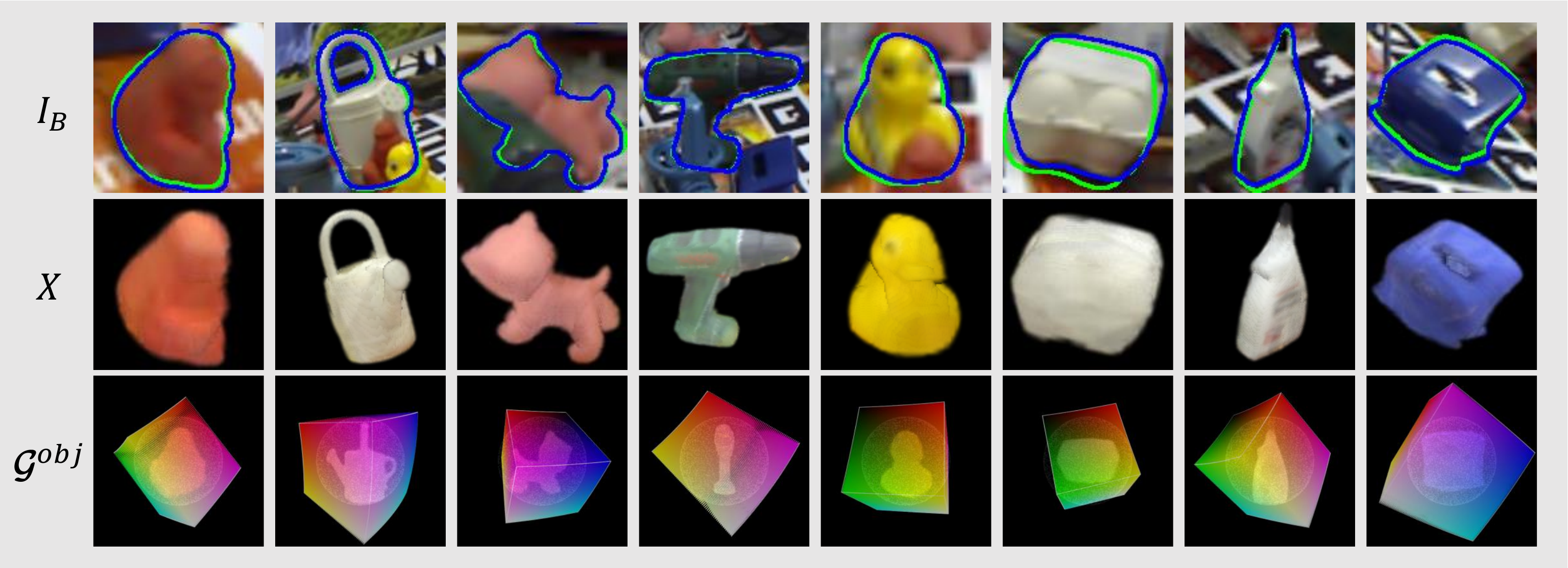}
\caption{\textbf{Qualitative results of LMO dataset.} The contours of projection by both label pose and predicted pose are represented as green and blue, respectively. The second row shows the projection by the estimated pose, and the third row shows the predicted grid, unit sphere, and the reference feature $\mathcal{F}$ in the object space.}
\label{fig:qualitative_LMO}
\end{figure}

Also, the accuracy of the LMO dataset is described in Table \ref{table:OCCLUSION}, and our method shows state-of-the-art performance. Some qualitative results on the LMO dataset are shown in Fig. \ref{fig:qualitative_LMO}. The figure shows that even the low-quality texture reference from the simple space-carving is enough for most samples. 

\begin{figure}[t]
\centering
\includegraphics[width=1.0\textwidth]{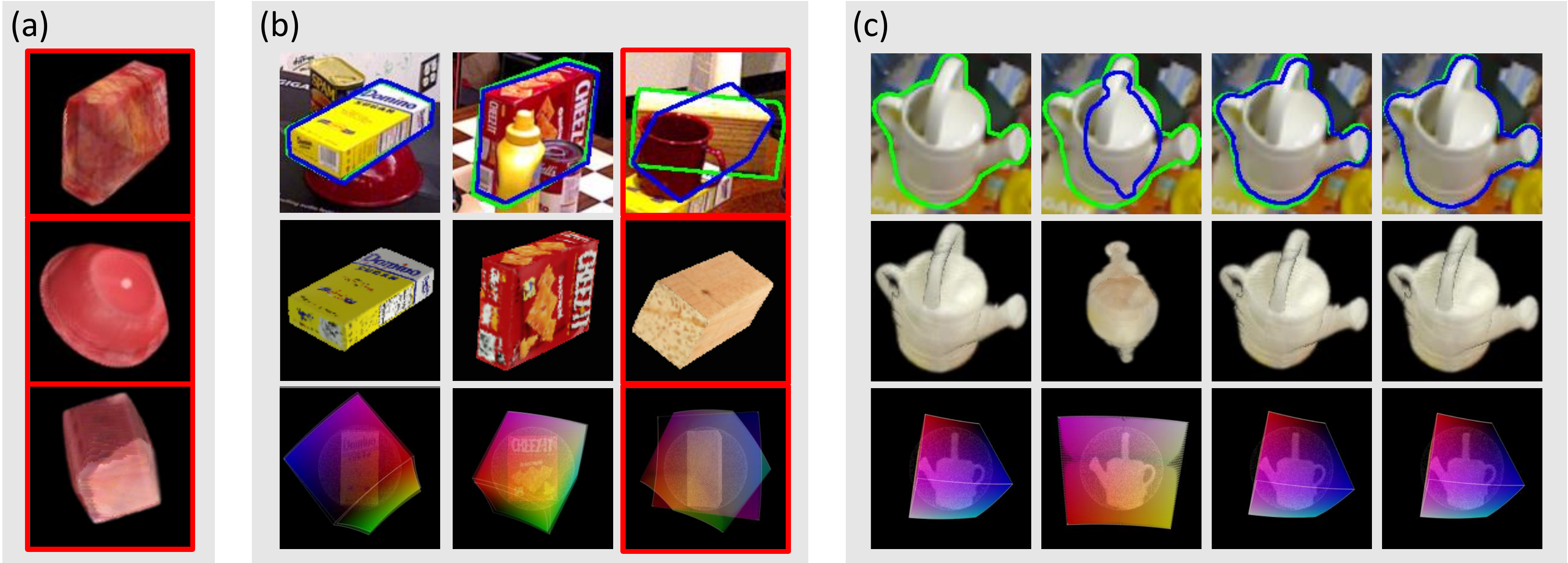}
\caption{\textbf{Qualitative results.} We visualize the contour of the projection by label pose in green, the prediction pose in blue, and error cases in a red box. (a) Space-carving failure cases in the YCBV dataset. (b) Qualitative results of the YCBV dataset. We demonstrate the rendered mesh images in the second row and superimposed $\mathcal{G}^{obj}$ of label and prediction in the third row. (c) Qualitative result of each iteration in the LM dataset. The first column shows the label, and each column from the second to the fourth indicates the initial pose and the predicted pose from each iteration.}
\label{fig:qualitative}
\end{figure}

However, we confirm that the reference feature-based projection is vulnerable to some conditions. For example, when the overall training view is similar, such as a video-based training set, or when majority of the samples are occluded, the shape of the reference quality is degraded. Because the YCBV has these characteristics, our reconstruction module fails to generate a plausible shape of the reference feature of some objects as shown in Fig. \ref{fig:qualitative}(a), therefore we used the mesh-based renderer in pytorch3d instead of our projector in the YCBV dataset. As stated in Table \ref{table:YCBV}, our model shows comparable performance to other state-of-the-art methods on the YCBV dataset. We also visualize some qualitative results of the YCBV dataset in Fig. \ref{fig:qualitative}(b). More qualitative results and comparisons with the other state-of-the-art methods are included in the supplementary material.

\begin{table}[t]
\caption{\textbf{Results on the LMO dataset.} The accuracy of baseline methods and our method in terms of the $ADD(\textbf{-}S)^{0.1d}$ score. Objects annotated with $(*)$ indicate symmetric pose ambiguity. A.O represents a pose estimator trained on all objects.}
{\resizebox{1.0\columnwidth}{!}
{\begin{tabular}{c|ccccccccc|cc}
\hline
\toprule
\multirow{2}{*}{Method}                                 & \begin{tabular}[c]{@{}c@{}}PoseCNN\\ \cite{Xiang2018posecnn}\end{tabular} & \begin{tabular}[c]{@{}c@{}}DeepIM\\ \cite{li2018deepim}\end{tabular} & \begin{tabular}[c]{@{}c@{}}PVNet\\ \cite{peng2019pvnet}\end{tabular} & \begin{tabular}[c]{@{}c@{}}S.Stage\\ \cite{hu2020single}\end{tabular} & \begin{tabular}[c]{@{}c@{}}HybridPose\\ \cite{song2020hybridpose}\end{tabular} & \begin{tabular}[c]{@{}c@{}}GDR-Net\\ \cite{wang2021gdr}\end{tabular} & \begin{tabular}[c]{@{}c@{}}GDR-Net\\ \cite{wang2021gdr}\end{tabular} & \begin{tabular}[c]{@{}c@{}}SO-Pose\\ \cite{di2021so}\end{tabular}  & \begin{tabular}[c]{@{}c@{}}RePOSE\\ \cite{iwase2021repose}\end{tabular} & DProST                                              & DProST                                              \\ \cline{2-12}
                                                       & \multicolumn{9}{c|}{w/ Mesh}                                                           & \multicolumn{2}{c}{w/o Mesh}                                                     \\ \hline
A.O                                                     & \checkmark                                                                & \checkmark                                                           &                                                                      &                                                                       &                                                                                &                                                                         & \checkmark                                                           & \checkmark                                                         &                                                                         &                                                     & \checkmark                                          \\ \hline
\begin{tabular}[c]{@{}c@{}}Training\\ Set\end{tabular} & \begin{tabular}[c]{@{}c@{}}real\\ +syn\end{tabular}                       & \begin{tabular}[c]{@{}c@{}}real\\ +syn\end{tabular}                  & \begin{tabular}[c]{@{}c@{}}real\\ +syn\end{tabular}                  & \begin{tabular}[c]{@{}c@{}}real\\ +syn\end{tabular}                        & \begin{tabular}[c]{@{}c@{}}real\\ +syn\end{tabular}                            & \begin{tabular}[c]{@{}c@{}}real\\ +pbr\end{tabular}               & \begin{tabular}[c]{@{}c@{}}real\\ +pbr\end{tabular}                  & \begin{tabular}[c]{@{}c@{}}real\\ +pbr\end{tabular}                  & \begin{tabular}[c]{@{}c@{}}real\\ +syn\end{tabular}                     & \begin{tabular}[c]{@{}c@{}}real\\ +pbr\end{tabular} & \begin{tabular}[c]{@{}c@{}}real\\ +pbr\end{tabular} \\ \hline
ape                                                     & 9.6                                                                       & 59.2                                                                 & 15.8                                                                 & 19.2                                                                       & 20.9                                                                           & 46.8                                                              & 44.9                                                                 & 48.4                                                                 & 31.1                                                                    & 50.9                                                & \textbf{51.4}                                                \\
can                                                     & 45.2                                                                      & 63.5                                                                 & 63.3                                                                 & 65.1                                                                       & 75.3                                                                           & \textbf{90.8}                                                                 & 79.7                                                                 & 85.8                                                              & 80.0                                                                    & 87.2                                                & 78.7                                                \\
cat                                                     & 0.9                                                                       & 26.2                                                                 & 16.7                                                                 & 18.9                                                                       & 24.9                                                                           & 40.5                                                                 & 30.6                                                                 & 32.7                                                              & 25.6                                                                    & 46.0                                                & \textbf{48.1}                                                \\
driller                                                 & 41.4                                                                      & 55.6                                                                 & 25.2                                                                 & 69.0                                                                       & 70.2                                                                           & 82.6                                                                 & 67.8                                                                 & 77.4                                                              & 73.1                                                                    & \textbf{86.1}                                                & 77.4                                                \\
duck                                                    & 19.6                                                                      & 52.4                                                                 & \textbf{65.7}                                                                 & 25.3                                                                       & 27.9                                                                           & 46.9                                                                 & 40.0                                                                 & 48.9                                                              & 43.0                                                                    & 47.7                                                & 45.4                                                \\
eggbox*                                                 & 22.0                                                                      & \textbf{63.0}                                                                 & 50.2                                                                 & 52.0                                                                       & 52.4                                                                           & 54.2                                                                 & 49.8                                                                 & 52.4                                                              & 51.7                                                                    & 46.9                                                & 55.3                                                \\
glue*                                                   & 38.5                                                                      & 71.7                                                                 & 49.6                                                                 & 51.4                                                                       & 53.8                                                                           & 75.8                                                                 & 73.7                                                                 & \textbf{78.3}                                                              & 54.3                                                                    & 68.5                                                & 76.9                                                \\ 
holepuncher                                             & 22.1                                                                      & 52.5                                                                 & 39.7                                                                 & 45.6                                                                       & 54.2                                                                           & 60.1                                                                 & 62.7                                                                 & \textbf{75.3}                                                              & 53.6                                                                    & 65.4                                                & 67.4                                                \\ \hline
Average                                                 & 24.9                                                                      & 55.5                                                                 & 40.8                                                                 & 43.3                                                                       & 47.5                                                                           & 62.2                                                                 & 56.1                                                                 & 62.4                                                              & 51.6                                                                    & 62.3                                                & \textbf{62.6}                                                \\ \bottomrule                          
\end{tabular}}}
\label{table:OCCLUSION}
\end{table}

\begin{table}[t]
\caption{\textbf{Results on the YCBV dataset.} The score of the baseline methods and our method in terms of $ADD(\textbf{-}S)^{0.1d}$ and $AUC$ of $ADD(\textbf{-}S)$. A.O represents a pose estimator trained on all objects.}
\centering
{\resizebox{1.0\columnwidth}{!}
{\begin{tabular}{c|ccccccc|cc}
\toprule
Method                    & \begin{tabular}[c]{@{}c@{}}PoseCNN\\ \cite{Xiang2018posecnn}\end{tabular} & \begin{tabular}[c]{@{}c@{}}PVNet\\ \cite{peng2019pvnet}\end{tabular} & \begin{tabular}[c]{@{}c@{}}DeepIM\\ \cite{li2018deepim}\end{tabular} & \begin{tabular}[c]{@{}c@{}}Cosypose\\ \cite{labbe2020cosypose}\end{tabular} &  \begin{tabular}[c]{@{}c@{}}GDR-Net\\ \cite{wang2021gdr}\end{tabular} & \begin{tabular}[c]{@{}c@{}}SO-Pose\\ \cite{di2021so}\end{tabular} & \begin{tabular}[c|]{@{}c@{}}RePOSE\\ \cite{iwase2021repose}\end{tabular} & DProST & DProST     \\ \hline
A.O                       & \checkmark                                                                &                                                                      & \checkmark                                                           & \checkmark                                                                  &                                                                       & \checkmark                                                        &                                                                               &        & \checkmark \\ \hline
$ADD(\textbf{-}S)^{0.1d}$        & 21.3                           & -                         & 53.6                      & -                                & 60.1                      & 56.8                 & 49.6                         & \textbf{65.1} &   43.8 \\
$AUC$ of $ADD(\textbf{-}S)$      & 61.3                           & 73.4                      & 81.9                      & \textbf{84.5}                    & 84.4                      & 83.9                 & 77.2                         & 77.4          &   69.2 \\ \bottomrule
\end{tabular}}}
\label{table:YCBV}

\end{table}

\subsection{Ablation Studies}
We conduct several ablation studies on the LM dataset. First, we test our method with the number of iterations as shown in Table \ref{table:ablation_estimator} $left$. The result shows that the performance is improved in the second iteration but saturates in more iterations. We visualize an example of iterative refinement in Fig. \ref{fig:qualitative}(c). 

We also visualize $\mathcal{G}^{obj}$s of the qualitative results in Fig. \ref{fig:qualitative}(b), (c). As shown in the figure, the difference of the grid area toward the camera and the opposite is caused by projective geometry. And since the ratio of two areas is inversely proportional to the object's distance, the shape of the $\mathcal{G}^{obj}$ reflects the object distance information. Consequently, as the GM loss leverages the shape of the $\mathcal{G}^{obj}$ to learn the distance, training even without GD loss is successful, as shown in Table \ref{table:ablation_estimator} $middle$. In other words, projective geometry helps our method to learn the distance to the object. We also confirm that $L^{GD}$ improves the performance, especially on the $ADD(\textbf{-}S)$ score, since it helps estimate more accurate distance of the grid in camera space. Additionally, the effect of GM loss compared to PM loss and image matching loss is reported in the supplementary material.

The performance about the number of points per ray ($N_{z}$) is in Table \ref{table:ablation_estimator} $right$, and $N_{z}=64$ shows the best in both $ADD(\textbf{-}S)$ and $Proj2D$ score.

Table \ref{table:ablation_ref} $left$ shows the experiment on the source of projection and view sampling method in reference feature. As demonstrated, a simple space-carving-based reference feature can replace the mesh without significant performance degradation, and FPS outperforms the random sampling of reference. 

Finally, we conduct an ablation study of the number of views ($N^{\mathcal{R}}$) for the reference image in Table \ref{table:ablation_ref} $right$. As shown in the table, eight reference images show the best performance. Note that because each voxel value of the reference feature is the average of the reference image, if there are too many references, the texture becomes blurry, and performance suffer. The quality of the reference feature depending on the number of views is demonstrated in the supplementary material.

\begin{table}[t]
\centering
\caption{\textbf{Ablation studies of the pose estimator.} $left$: Ablation of the number of iterations. $middle$: Ablation of the loss function. $right$: Ablation of the $N_z$.}
{\resizebox{0.30\columnwidth}{!}
{\begin{tabular}{c|ccc|cc}
\toprule
\multirow{2}{*}{Iter} & \multicolumn{3}{c|}{$ADD(\textbf{-}S)$}& \multicolumn{2}{c}{\textit{$Proj2D$}} \\
                           & 0.02d   & 0.05d   & 0.10d   & 2pix    & 5pix \\ \hline
1                          & 22.6    & 68.9    & 93.2    & 62.4    & 97.6 \\
2                          & 48.1    & 85.8    & \textbf{97.7}    & 89.3    & \textbf{99.2} \\
3                          & \textbf{50.8}    & \textbf{86.3}    & \textbf{97.7}    & \textbf{90.8}    & 99.1 \\ \bottomrule
\end{tabular}}} 
{\resizebox{0.38\columnwidth}{!}
{\begin{tabular}{c|ccc|cc}
\toprule
\multirow{2}{*}{Loss}               & \multicolumn{3}{c|}{$ADD(\textbf{-}S)$}& \multicolumn{2}{c}{\textit{$Proj2D$}} \\
                                    & 0.02d   & 0.05d   & 0.10d   & 2pix   & 5pix \\ \hline
                                    &         &         &         &        &      \\ [-0.6em]
$\mathcal{L}^{GM}$                  & 35.5    & 75.8    & 94.5    & 86.6   & 99.0 \\
                                    &         &         &         &        &      \\ [-0.6em]
$\mathcal{L}^{GM}+\mathcal{L}^{GD}$ & \textbf{48.1}    & \textbf{85.8}    & \textbf{97.7}    & \textbf{89.3}   & \textbf{99.2} \\\bottomrule
\end{tabular}}} 
{\resizebox{0.30\columnwidth}{!}
{\begin{tabular}{c|ccc|cc}
\toprule
\multirow{2}{*}{$N_z$}               & \multicolumn{3}{c|}{$ADD(\textbf{-}S)$}& \multicolumn{2}{c}{\textit{$Proj2D$}} \\
                    & 0.02d   & 0.05d   & 0.10d   & 2pix   & 5pix \\ \hline
32                  & 46.3    & 85.4    & 97.6    & 88.0   & 99.1 \\
64                  & \textbf{48.1}    & \textbf{85.8}    & \textbf{97.7}    & \textbf{89.3}   & \textbf{99.2} \\ 
128                 & 47.1    & 85.4    & 97.5    & 88.8   & 99.1 \\ \bottomrule
\end{tabular}}}
\label{table:ablation_estimator}
\end{table}

\begin{table}[t]
\centering
\caption{\textbf{Ablation studies of the 3D feature.} $left$: Ablation of the projection source and reference generation method. $right$: Ablation of the number of references.}
{\resizebox{0.47\columnwidth}{!}
{\begin{tabular}{cc|ccc|cc}
\toprule
\multirow{2}{*}{Source} & \multirow{2}{*}{Sampling} & \multicolumn{3}{c|}{$ADD(\textbf{-}S)$} & \multicolumn{2}{c}{\textit{$Proj2D$}}\\ 
                        &                           & 0.02d   & 0.05d   & 0.10d   & 2pix      & 5pix \\ \hline
mesh                    &                           & \textbf{51.6}    & \textbf{87.8}    & \textbf{98.3}    & \textbf{91.5}      & 99.1 \\ \hline
reference               & random                    & 45.9    & 84.7    & 97.5    & 88.8      & 99.1 \\
reference               & FPS                       & 48.1    & 85.8    & 97.7    & 89.3      & \textbf{99.2} \\ \bottomrule
\end{tabular}}}
\qquad
{\resizebox{0.32\columnwidth}{!}
{\begin{tabular}{c|ccc|cc}
\toprule
\multirow{2}{*}{$N^{\mathcal{R}}$} & \multicolumn{3}{c|}{$ADD(\textbf{-}S)$}& \multicolumn{2}{c}{\textit{$Proj2D$}}  \\
                         & 0.02d   & 0.05d   & 0.10d   & 2pix   & 5pix \\ \hline
4                        & 44.1    & 84.2    & 97.5    & 87.5   & 98.9 \\
8                        & \textbf{48.1}    & \textbf{85.8}    & \textbf{97.7}    & \textbf{89.3}   & \textbf{99.2} \\
16                       & 45.6    & 84.9    & 97.6    & 87.8   & 99.1 \\ \bottomrule
\end{tabular}}}
\label{table:ablation_ref}
\end{table}

\subsection{Runtime Analysis}
The experiments is conducted on an AMD Ryzen 9 3900X 12-Core CPU with an NVIDIA Geforce RTX 2080Ti GPU. In addition to the 1.21 s for space carving and 15 ms in detection, our method takes 0.22 ms in grid generator, 2.06ms in the projector, and 3.86 ms in the pose estimator for each step. Note that, since our method focuses on the seen object scenario, the reference feature is created only once for each object in training phase and used repeatedly without additional generation. Because the grid from the grid generator is directly used as a sampling grid, our projector is faster than the mesh rendering function in pytorch3d, which takes 3.14 ms.

\section{Conclusion}
We have proposed a new 6D object pose estimation method based on a grid of the object space. To accomplish this, we have designed the DProST model that elaborately considers the projective geometry, while reducing the number of computations by focusing the grid on object space RoI. Additionally, we have introduced new objective functions, GM and GD, which can be used to train the pose estimator based on the object space grid. We also have proposed a simple space-carving-based reference feature generation method, which can replace the mesh data in the projection stage. Experiments have shown that DProST outperforms the state-of-the-art pose estimation method even without mesh data on the LM and LMO datasets and shows competitive performance on the YCBV dataset with mesh data. We plan to apply other 3D reconstruction methods based on deep learning to the DProST in the future. Also, applying our method to unseen or categorical objects would be one area for future research.

\subsubsection{Acknowledgement}
This research was supported in part by LG AI Research, in part by Institute of Information \& 
Communications Technology Planning \& 
Evaluation (IITP) Grant funded by the Korea government(MSIT) [NO.2021-0-01343, Artificial Intelligence Graduate School Program (Seoul National University)],
and partially by the BK21 FOUR program of the Education and Research Program for Future ICT Pioneers, Seoul National University in 2022.

%
%
\bibliographystyle{splncs04}
\bibliography{egbib}

\begin{thebibliography}{10}
\providecommand{\url}[1]{\texttt{#1}}
\providecommand{\urlprefix}{URL }
\providecommand{\doi}[1]{https://doi.org/#1}

\bibitem{brachmann2014learning}
Brachmann, E., Krull, A., Michel, F., Gumhold, S., Shotton, J., Rother, C.:
  Learning 6d object pose estimation using 3d object coordinates. In: European
  conference on computer vision. pp. 536--551. Springer (2014)

\bibitem{brachmann2016uncertainty}
Brachmann, E., Michel, F., Krull, A., Yang, M.Y., Gumhold, S., et~al.:
  Uncertainty-driven 6d pose estimation of objects and scenes from a single rgb
  image. In: Proceedings of the IEEE conference on computer vision and pattern
  recognition. pp. 3364--3372 (2016)

\bibitem{cai2018cascade}
Cai, Z., Vasconcelos, N.: Cascade r-cnn: Delving into high quality object
  detection. In: Proceedings of the IEEE conference on computer vision and
  pattern recognition. pp. 6154--6162 (2018)

\bibitem{chen2017multi}
Chen, X., Ma, H., Wan, J., Li, B., Xia, T.: Multi-view 3d object detection
  network for autonomous driving. In: Proceedings of the IEEE conference on
  Computer Vision and Pattern Recognition. pp. 1907--1915 (2017)

\bibitem{chen2020category}
Chen, X., Dong, Z., Song, J., Geiger, A., Hilliges, O.: Category level object
  pose estimation via neural analysis-by-synthesis. In: European Conference on
  Computer Vision. pp. 139--156. Springer (2020)

\bibitem{cheng20216d}
Cheng, Y., Zhu, H., Sun, Y., Acar, C., Jing, W., Wu, Y., Li, L., Tan, C., Lim,
  J.H.: 6d pose estimation with correlation fusion. In: 2020 25th International
  Conference on Pattern Recognition (ICPR). pp. 2988--2994. IEEE (2021)

\bibitem{deng2009imagenet}
Deng, J., Dong, W., Socher, R., Li, L.J., Li, K., Fei-Fei, L.: Imagenet: A
  large-scale hierarchical image database. In: 2009 IEEE conference on computer
  vision and pattern recognition. pp. 248--255. Ieee (2009)

\bibitem{di2021so}
Di, Y., Manhardt, F., Wang, G., Ji, X., Navab, N., Tombari, F.: So-pose:
  Exploiting self-occlusion for direct 6d pose estimation. In: Proceedings of
  the IEEE/CVF International Conference on Computer Vision. pp. 12396--12405
  (2021)

\bibitem{gao2020generalizing}
Gao, C., Liu, X., Gu, W., Killeen, B., Armand, M., Taylor, R., Unberath, M.:
  Generalizing spatial transformers to projective geometry with applications to
  2d/3d registration. In: International Conference on Medical Image Computing
  and Computer-Assisted Intervention. pp. 329--339. Springer (2020)

\bibitem{he2017mask}
He, K., Gkioxari, G., Doll{\'a}r, P., Girshick, R.: Mask r-cnn. In: Proceedings
  of the IEEE international conference on computer vision. pp. 2961--2969
  (2017)

\bibitem{he2016deep}
He, K., Zhang, X., Ren, S., Sun, J.: Deep residual learning for image
  recognition. In: Proceedings of the IEEE conference on computer vision and
  pattern recognition. pp. 770--778 (2016)

\bibitem{hinterstoisser2012model}
Hinterstoisser, S., Lepetit, V., Ilic, S., Holzer, S., Bradski, G., Konolige,
  K., Navab, N.: Model based training, detection and pose estimation of
  texture-less 3d objects in heavily cluttered scenes. In: Asian conference on
  computer vision. pp. 548--562. Springer (2012)

\bibitem{hodavn2020bop}
Hoda{\v{n}}, T., Sundermeyer, M., Drost, B., Labb{\'e}, Y., Brachmann, E.,
  Michel, F., Rother, C., Matas, J.: Bop challenge 2020 on 6d object
  localization. In: European Conference on Computer Vision. pp. 577--594.
  Springer (2020)

\bibitem{hu2020single}
Hu, Y., Fua, P., Wang, W., Salzmann, M.: Single-stage 6d object pose
  estimation. In: Proceedings of the IEEE/CVF conference on computer vision and
  pattern recognition. pp. 2930--2939 (2020)

\bibitem{hu2019segmentation}
Hu, Y., Hugonot, J., Fua, P., Salzmann, M.: Segmentation-driven 6d object pose
  estimation. In: Proceedings of the IEEE/CVF Conference on Computer Vision and
  Pattern Recognition. pp. 3385--3394 (2019)

\bibitem{iwase2021repose}
Iwase, S., Liu, X., Khirodkar, R., Yokota, R., Kitani, K.M.: Repose: Fast 6d
  object pose refinement via deep texture rendering. In: Proceedings of the
  IEEE/CVF International Conference on Computer Vision. pp. 3303--3312 (2021)

\bibitem{jaderberg2015spatial}
Jaderberg, M., Simonyan, K., Zisserman, A., et~al.: Spatial transformer
  networks. Advances in neural information processing systems  \textbf{28},
  2017--2025 (2015)

\bibitem{kingma2014adam}
Kingma, D.P., Ba, J.: Adam: A method for stochastic optimization. arXiv
  preprint arXiv:1412.6980  (2014)

\bibitem{labbe2020cosypose}
Labb{\'e}, Y., Carpentier, J., Aubry, M., Sivic, J.: Cosypose: Consistent
  multi-view multi-object 6d pose estimation. In: European Conference on
  Computer Vision. pp. 574--591. Springer (2020)

\bibitem{lepetit2009epnp}
Lepetit, V., Moreno-Noguer, F., Fua, P.: Epnp: An accurate o (n) solution to
  the pnp problem. International journal of computer vision  \textbf{81}(2),
  ~155 (2009)

\bibitem{li2018deepim}
Li, Y., Wang, G., Ji, X., Xiang, Y., Fox, D.: Deepim: Deep iterative matching
  for 6d pose estimation. In: Proceedings of the European Conference on
  Computer Vision (ECCV). pp. 683--698 (2018)

\bibitem{li2019cdpn}
Li, Z., Wang, G., Ji, X.: Cdpn: Coordinates-based disentangled pose network for
  real-time rgb-based 6-dof object pose estimation. In: Proceedings of the
  IEEE/CVF International Conference on Computer Vision. pp. 7678--7687 (2019)

\bibitem{marchand2015pose}
Marchand, E., Uchiyama, H., Spindler, F.: Pose estimation for augmented
  reality: a hands-on survey. IEEE transactions on visualization and computer
  graphics  \textbf{22}(12),  2633--2651 (2015)

\bibitem{mildenhall2020nerf}
Mildenhall, B., Srinivasan, P.P., Tancik, M., Barron, J.T., Ramamoorthi, R.,
  Ng, R.: Nerf: Representing scenes as neural radiance fields for view
  synthesis. In: European conference on computer vision. pp. 405--421. Springer
  (2020)

\bibitem{RenderNet2018}
Nguyen-Phuoc, T., Li, C., Balaban, S., Yang, Y.L.: Rendernet: A deep
  convolutional network for differentiable rendering from 3d shapes. In:
  Advances in Neural Information Processing Systems 31 (2018)

\bibitem{oberweger2018making}
Oberweger, M., Rad, M., Lepetit, V.: Making deep heatmaps robust to partial
  occlusions for 3d object pose estimation. In: Proceedings of the European
  Conference on Computer Vision (ECCV). pp. 119--134 (2018)

\bibitem{park2020latentfusion}
Park, K., Mousavian, A., Xiang, Y., Fox, D.: Latentfusion: End-to-end
  differentiable reconstruction and rendering for unseen object pose
  estimation. In: Proceedings of the IEEE/CVF conference on computer vision and
  pattern recognition. pp. 10710--10719 (2020)

\bibitem{park2019pix2pose}
Park, K., Patten, T., Vincze, M.: Pix2pose: Pixel-wise coordinate regression of
  objects for 6d pose estimation. In: Proceedings of the IEEE/CVF International
  Conference on Computer Vision. pp. 7668--7677 (2019)

\bibitem{park2020neural}
Park, K., Patten, T., Vincze, M.: Neural object learning for 6d pose estimation
  using a few cluttered images. In: European Conference on Computer Vision. pp.
  656--673. Springer (2020)

\bibitem{paszke2019pytorch}
Paszke, A., Gross, S., Massa, F., Lerer, A., Bradbury, J., Chanan, G., Killeen,
  T., Lin, Z., Gimelshein, N., Antiga, L., et~al.: Pytorch: An imperative
  style, high-performance deep learning library. Advances in neural information
  processing systems  \textbf{32},  8026--8037 (2019)

\bibitem{peng2019pvnet}
Peng, S., Liu, Y., Huang, Q., Zhou, X., Bao, H.: Pvnet: Pixel-wise voting
  network for 6dof pose estimation. In: Proceedings of the IEEE/CVF Conference
  on Computer Vision and Pattern Recognition. pp. 4561--4570 (2019)

\bibitem{rad2017bb8}
Rad, M., Lepetit, V.: Bb8: A scalable, accurate, robust to partial occlusion
  method for predicting the 3d poses of challenging objects without using
  depth. In: Proceedings of the IEEE International Conference on Computer
  Vision. pp. 3828--3836 (2017)

\bibitem{ravi2020accelerating}
Ravi, N., Reizenstein, J., Novotny, D., Gordon, T., Lo, W.Y., Johnson, J.,
  Gkioxari, G.: Accelerating 3d deep learning with pytorch3d. arXiv preprint
  arXiv:2007.08501  (2020)

\bibitem{redmon2018yolov3}
Redmon, J., Farhadi, A.: Yolov3: An incremental improvement. arXiv preprint
  arXiv:1804.02767  (2018)

\bibitem{ren2015faster}
Ren, S., He, K., Girshick, R., Sun, J.: Faster r-cnn: Towards real-time object
  detection with region proposal networks. Advances in neural information
  processing systems  \textbf{28},  91--99 (2015)

\bibitem{sitzmann2019deepvoxels}
Sitzmann, V., Thies, J., Heide, F., Nie{\ss}ner, M., Wetzstein, G., Zollhofer,
  M.: Deepvoxels: Learning persistent 3d feature embeddings. In: Proceedings of
  the IEEE/CVF Conference on Computer Vision and Pattern Recognition. pp.
  2437--2446 (2019)

\bibitem{song2020hybridpose}
Song, C., Song, J., Huang, Q.: Hybridpose: 6d object pose estimation under
  hybrid representations. In: Proceedings of the IEEE/CVF conference on
  computer vision and pattern recognition. pp. 431--440 (2020)

\bibitem{tremblay2018deep}
Tremblay, J., To, T., Sundaralingam, B., Xiang, Y., Fox, D., Birchfield, S.:
  Deep object pose estimation for semantic robotic grasping of household
  objects. In: Conference on Robot Learning. pp. 306--316. PMLR (2018)

\bibitem{wang2019densefusion}
Wang, C., Xu, D., Zhu, Y., Mart{\'\i}n-Mart{\'\i}n, R., Lu, C., Fei-Fei, L.,
  Savarese, S.: Densefusion: 6d object pose estimation by iterative dense
  fusion. In: Proceedings of the IEEE/CVF conference on computer vision and
  pattern recognition. pp. 3343--3352 (2019)

\bibitem{wang2021gdr}
Wang, G., Manhardt, F., Tombari, F., Ji, X.: Gdr-net: Geometry-guided direct
  regression network for monocular 6d object pose estimation. In: Proceedings
  of the IEEE/CVF Conference on Computer Vision and Pattern Recognition. pp.
  16611--16621 (2021)

\bibitem{wang2019normalized}
Wang, H., Sridhar, S., Huang, J., Valentin, J., Song, S., Guibas, L.J.:
  Normalized object coordinate space for category-level 6d object pose and size
  estimation. In: Proceedings of the IEEE/CVF Conference on Computer Vision and
  Pattern Recognition. pp. 2642--2651 (2019)

\bibitem{Xiang2018posecnn}
Xiang, Y., Schmidt, T., Narayanan, V., Fox, D.: Posecnn: A convolutional neural
  network for 6d object pose estimation in cluttered scenes. In: Proceedings of
  Robotics: Science and Systems. Pittsburgh, Pennsylvania (June 2018).
  \doi{10.15607/RSS.2018.XIV.019}

\bibitem{xu2018pointfusion}
Xu, D., Anguelov, D., Jain, A.: Pointfusion: Deep sensor fusion for 3d bounding
  box estimation. In: Proceedings of the IEEE conference on computer vision and
  pattern recognition. pp. 244--253 (2018)

\bibitem{yen2021inerf}
Yen-Chen, L., Florence, P., Barron, J.T., Rodriguez, A., Isola, P., Lin, T.Y.:
  inerf: Inverting neural radiance fields for pose estimation. In: 2021
  IEEE/RSJ International Conference on Intelligent Robots and Systems (IROS).
  pp. 1323--1330. IEEE (2021)

\bibitem{zakharov2019dpod}
Zakharov, S., Shugurov, I., Ilic, S.: Dpod: 6d pose object detector and
  refiner. In: Proceedings of the IEEE/CVF International Conference on Computer
  Vision. pp. 1941--1950 (2019)

\bibitem{zhou2019continuity}
Zhou, Y., Barnes, C., Lu, J., Yang, J., Li, H.: On the continuity of rotation
  representations in neural networks. In: Proceedings of the IEEE/CVF
  Conference on Computer Vision and Pattern Recognition. pp. 5745--5753 (2019)

\bibitem{zhu2014single}
Zhu, M., Derpanis, K.G., Yang, Y., Brahmbhatt, S., Zhang, M., Phillips, C.,
  Lecce, M., Daniilidis, K.: Single image 3d object detection and pose
  estimation for grasping. In: 2014 IEEE International Conference on Robotics
  and Automation (ICRA). pp. 3936--3943. IEEE (2014)

\end{thebibliography}

\title{Supplementary Material - DProST: Dynamic Projective Spatial Transformer Network for \\ 6D Pose Estimation} 

\titlerunning{DProST: Dynamic Projective Spatial Transformer Network}
\author{Jaewoo Park\inst{1,2}\orcidlink{0000-0002-6816-4381} \and
Nam Ik Cho\inst{1,2,3}\orcidlink{0000-0001-5297-4649}}
\authorrunning{J. Park et al.}
%
\institute{Department of ECE \& INMC, Seoul National University, Seoul, Korea \and 
SNU-LG AI Research Center, Seoul, Korea \and
IPAI, Seoul National University, Seoul, Korea \\
\email{\{bjw0611,nicho\}@snu.ac.kr}}

\maketitle

\appendix
\renewcommand\thefigure{\thesection.\arabic{figure}}  
\renewcommand\thetable{\thesection.\arabic{table}} 

\section{Details of Pose Estimator}
The disentangled representation is used for the pose estimator output as in \cite{labbe2020cosypose}. For each iteration ${i}$, our ResNet34 \cite{he2016deep} based pose estimator predicts relative translation on image space, which can be written as follows:
\begin{align}
v_x^i & = f_x\left(\frac{t_x^{i}}{t_z^{i}} - \frac{t_x^{i-1}}{t_z^{i-1}}\right) \label{equ:translation_x}, \\
v_y^i & = f_y\left(\frac{t_y^{i}}{t_z^{i}} - \frac{t_y^{i-1}}{t_z^{i-1}}\right) \label{equ:translation_y}, \\
v_z^i & = \frac{t_z^{i}}{t_z^{i-1}} \label{equ:translation_z},
\end{align}
where $v_x^i$ and $v_y^i$ are pixel-wise translation estimation, $v_z^i$ is the relative scale of the object, $f_x$ and $f_y$ are focal lengths of $x$ and $y$ axis in intrinsic matrix, respectively, and $t_x^i$, $t_y^i$, and $t_z^i$ are the components of the translation vector $t_i$. With the Equations (\ref{equ:translation_x}), (\ref{equ:translation_y}), and (\ref{equ:translation_z}), we update the translation vector $t_{i-1}$ to $t_i$ for each iteration.

The network also predicts the two three-dimensional vectors $e_1^i$ and $e_2^i$ for rotation representation as in \cite{labbe2020cosypose,zhou2019continuity}, which can be converted to relative rotation matrix as follows:
\begin{align}
r_1^i & = \frac{e_1^i}{\|e_1^i\|_2}, \\
r_3^i & = \frac{r_1^i \land e_2^i}{\|e_2^i\|_2}, \\
r_2^i & = r_3^i \land r_1^i,
\label{equ:rotation} 
\end{align}
where $\land$ represents the cross product, and $r_1^i$, $r_2^i$, and $r_3^i$ are the vectors in the relative rotation matrix of the $i$-th iteration.

To localize the initial projection of object in bounding box $(x, y, w, h)$ as in \cite{labbe2020cosypose}, we set the initial translation $t_0$ as follows:
\begin{align}
t_x^0 & = (c_x - p_x)\frac{t_z^0}{f_x} \\
t_y^0 & = (c_y - p_y)\frac{t_z^0}{f_y} \\
t_z^0 & = \frac{d}{2}(\frac{f_x}{w} + \frac{f_y}{h})
\label{equ:initialize} 
\end{align}
where $t_x^0$, $t_y^0$, and $t_z^0$ are components of initial translation vector, $c_x$ and $c_y$ are center of the bounding box, $p_x$ and $p_y$ are principal point, and $d$ represents the diameter of object which is set to two in our normalized object setting. Additionally, we use the identity matrix for the initial rotation matrix $R_0$.

\section{Additional Ablation Studies}
\begin{figure}
\centering
\includegraphics[width=1.0\columnwidth]{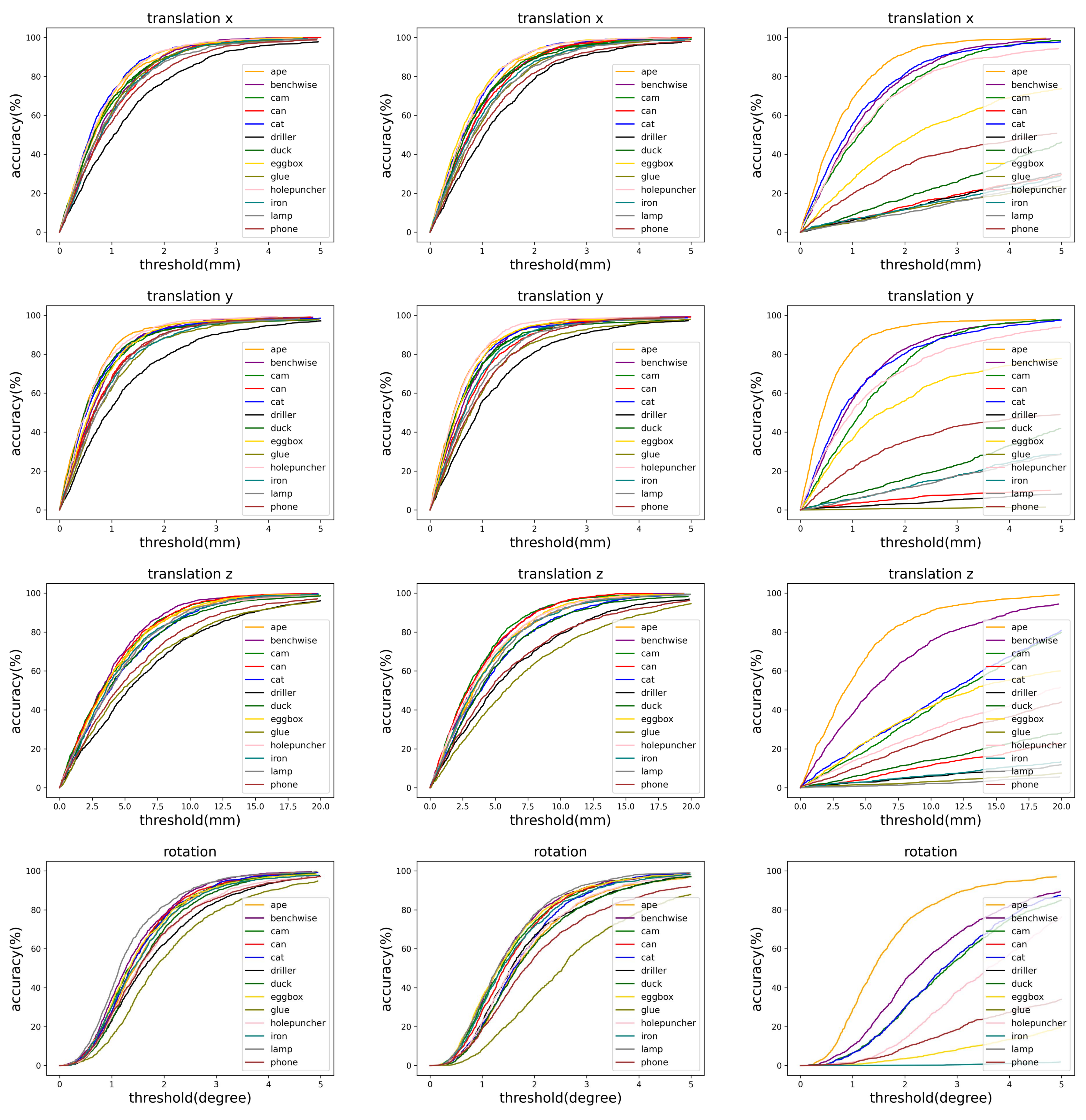}
\caption{\textbf{Comparison of accuracies according to loss functions on the LM dataset}: Each column demonstrates the accuracy of GM loss, PM loss, and IM loss on the LM dataset, respectively. We visualize the translation error for each axis in the first three rows and rotation error in the last row.}
\label{fig:error_graph}
\end{figure}

The performances of grid matching (GM), point matching (PM), and image matching (IM) objective function are visualized in Fig. \ref{fig:error_graph}. The mean distance between object vertices in camera space is used for PM as in \cite{labbe2020cosypose,li2018deepim}, and mean-squared-error between projection in image space is used for IM as \cite{chen2020category}. We confirm that the GM predict the rotation more accurately than PM in long objects such as glue in the LM dataset, and the shape bias in PM causes performance degradation. In detail, since the PM uses the object vertices, even if the actual rotation error of two predictions are same, the loss vary depending on the direction of misalignment. For example, the PM loss of the axial rotation error of a long object is relatively lower than other direction errors, which hinders accurate prediction of axial rotation. On the other hand, since the GM uses a uniformly distributed grid, no performance degradation due to shape bias has occurred. Furthermore, since the GM reflects the projective geometry on the grid shape and leverages it to predict the distance to object, the GM shows better results than PM in z-axis translation. Unlike the GM and PM, since the IM does not leverage the 3D location information, IM based model fails to predict the pose in some objects and shows low performance.

\section{Additional Qualitative Results}
\begin{figure}
\centering
\includegraphics[width=1.0\columnwidth]{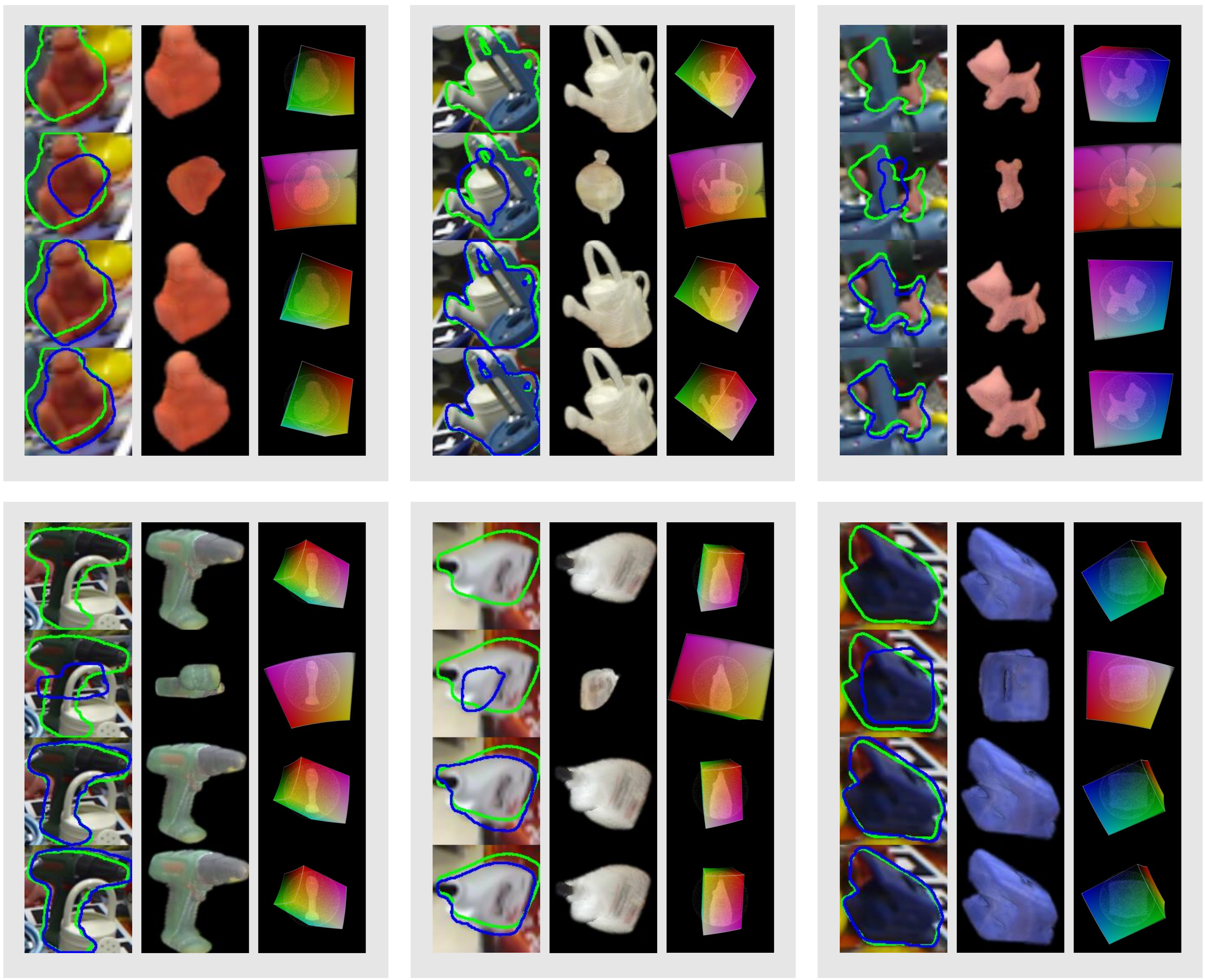}
\caption{\textbf{Qualitative results of iteration on the LMO dataset}: For each column in the grey cell, we visualize the contours of the projection, projection by estimated pose, and object space grid. From the first row to the fourth row in each gray cell, the ground-truth pose, the initial pose, the first iteration pose, and the second iteration pose are visualized. The predicted and ground-truth poses are represented by blue and green contour, respectively.}
\label{fig:LMO_iter}
\end{figure}

\begin{figure}
\centering
\includegraphics[width=1.0\columnwidth]{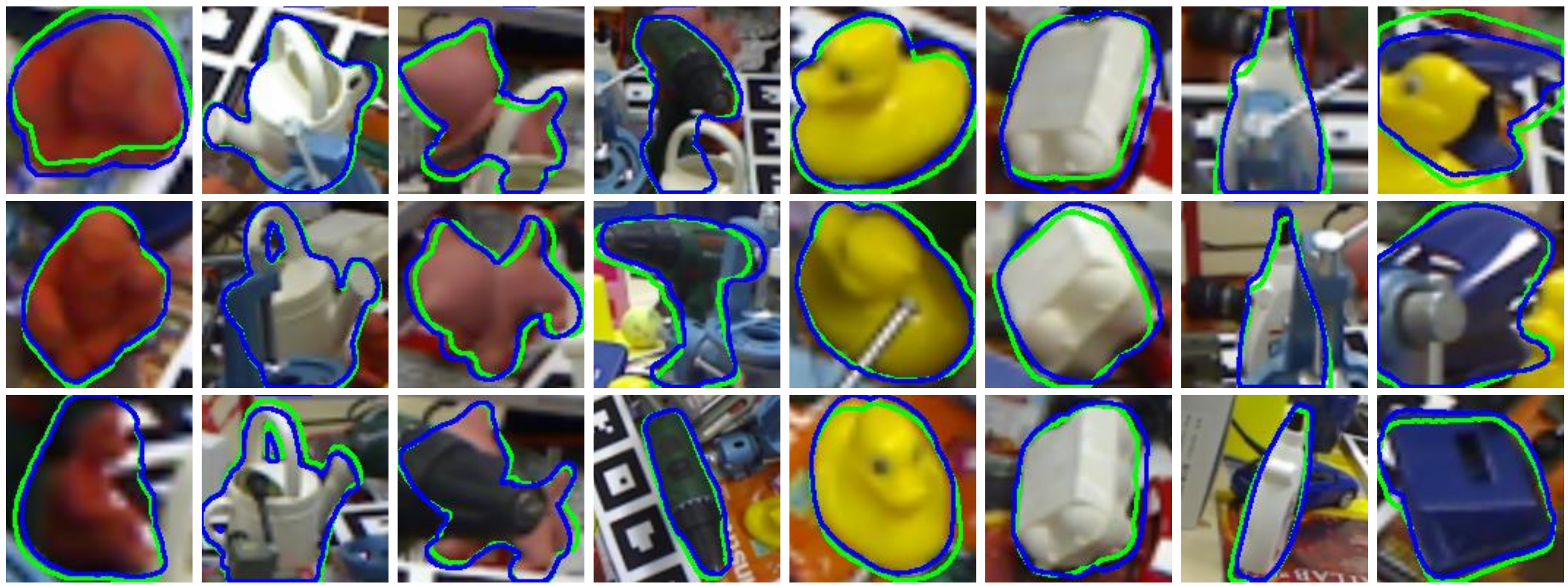}
\caption{\textbf{Qualitative results on the LMO dataset}: We visualize additional results on the LMO dataset. The predicted pose's contour is represented by blue, and the ground truth pose's contour is demonstrated by green.}
\label{fig:LMO_example}
\end{figure}

\begin{figure}
\centering
\includegraphics[width=1.0\columnwidth]{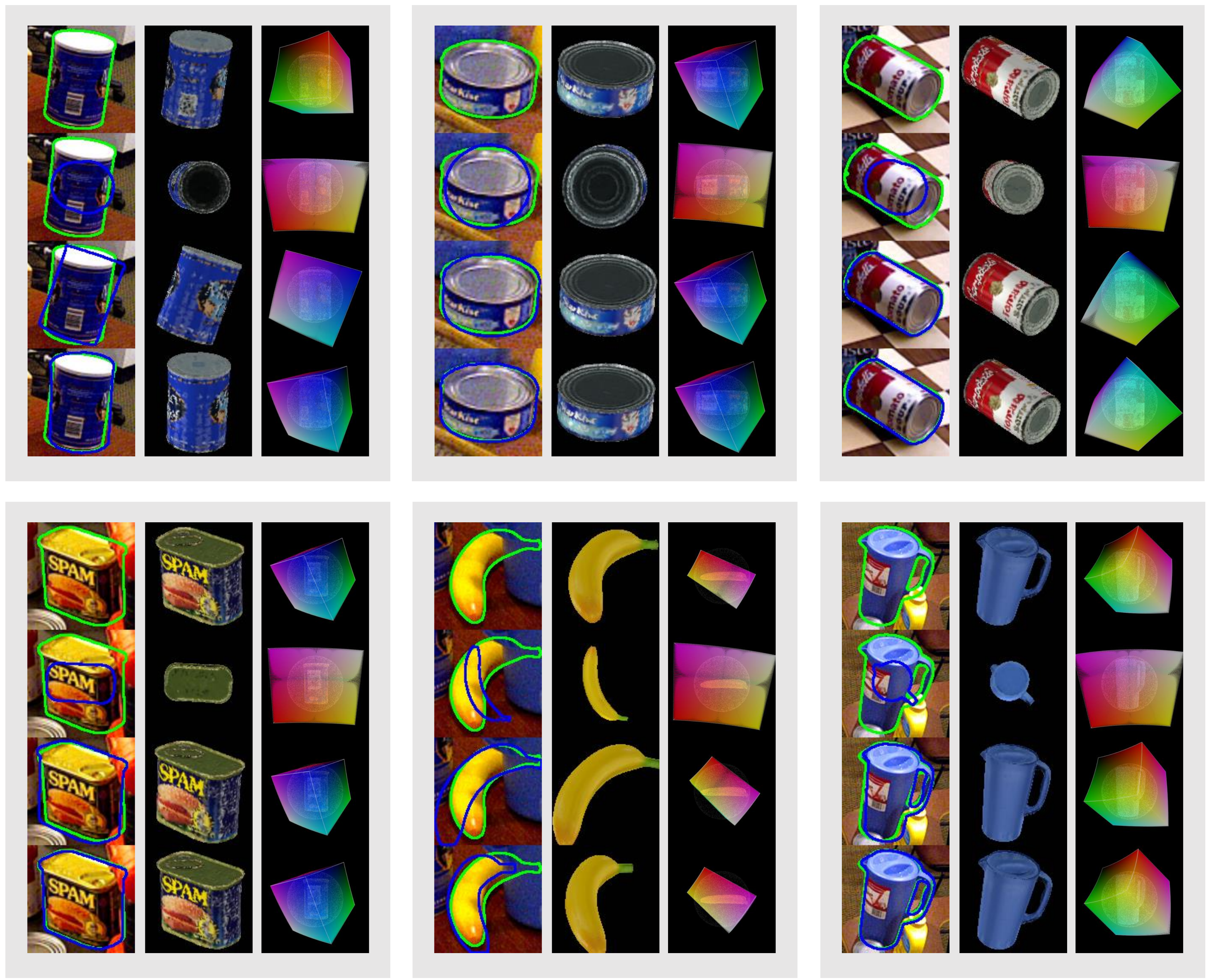}
\caption{\textbf{Qualitative results of iteration on the YCBV dataset}: For each column in grey cells, we visualize the contours of the projection, projection results, and object space grid. From the first row to the fourth row in each gray cell, the ground-truth pose, the initial pose, the first iteration pose, and the second iteration pose are visualized, respectively. The predicted pose's contour is represented by blue, and the ground truth pose's contour is demonstrated by green.}
\label{fig:YCBV_iter}
\end{figure}

\begin{figure}
\centering
\includegraphics[width=1.0\columnwidth]{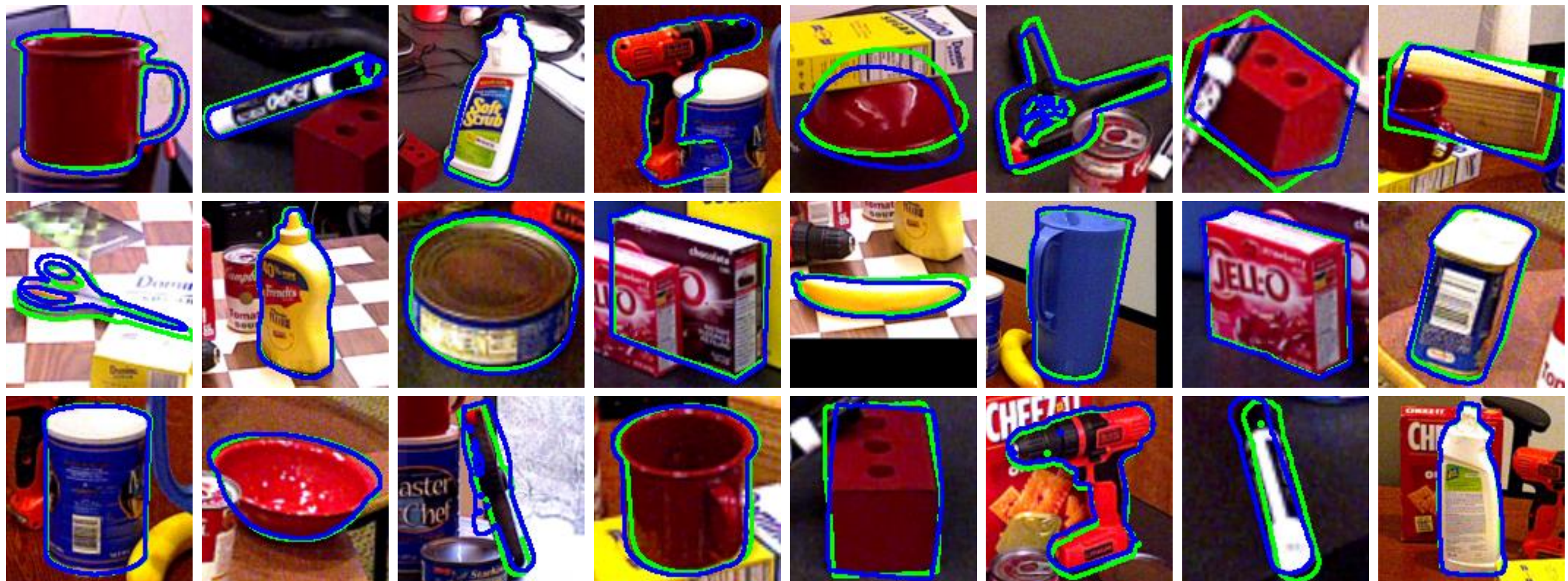}
\caption{\textbf{Qualitative results on the YCBV dataset}: We visualize additional results on the YCBV dataset. The predicted pose's contour is represented by blue, and the ground truth pose's contour is demonstrated by green.}
\label{fig:YCBV_example}
\end{figure}

\begin{figure}
\centering
\includegraphics[width=1.0\columnwidth]{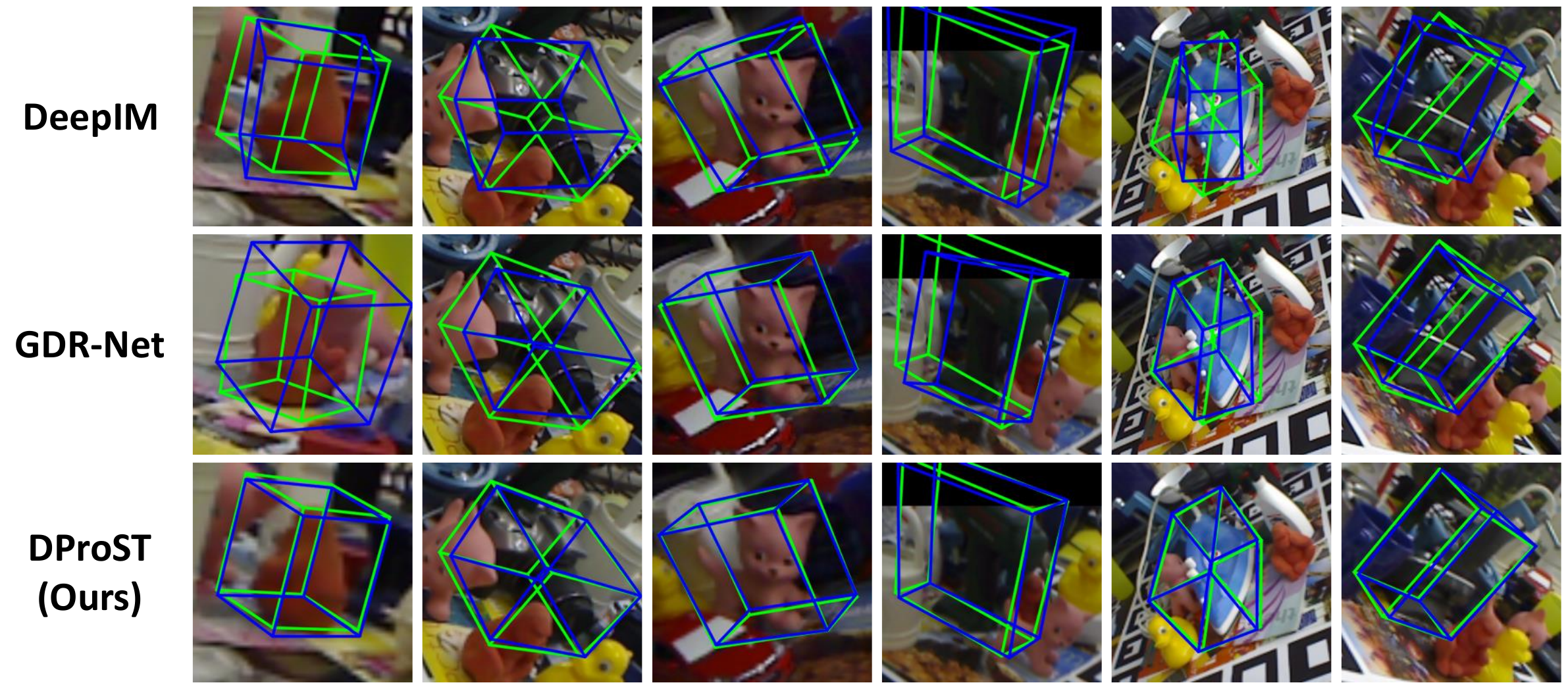}
\caption{\textbf{Comparison of qualitative results with other SOTA methods}: For each row, we visualize the qualitative results of DeepIM \cite{li2018deepim} (first row), GDR-Net \cite{wang2021gdr} (second row), and our method (last row) on the LM dataset. The green and blue boxes visualize the projection of the object 3D bounding box using ground-truth and predicted pose, respectively.}
\label{fig:qualitative_other}
\end{figure}

\begin{figure}
\centering
\includegraphics[width=1.0\columnwidth]{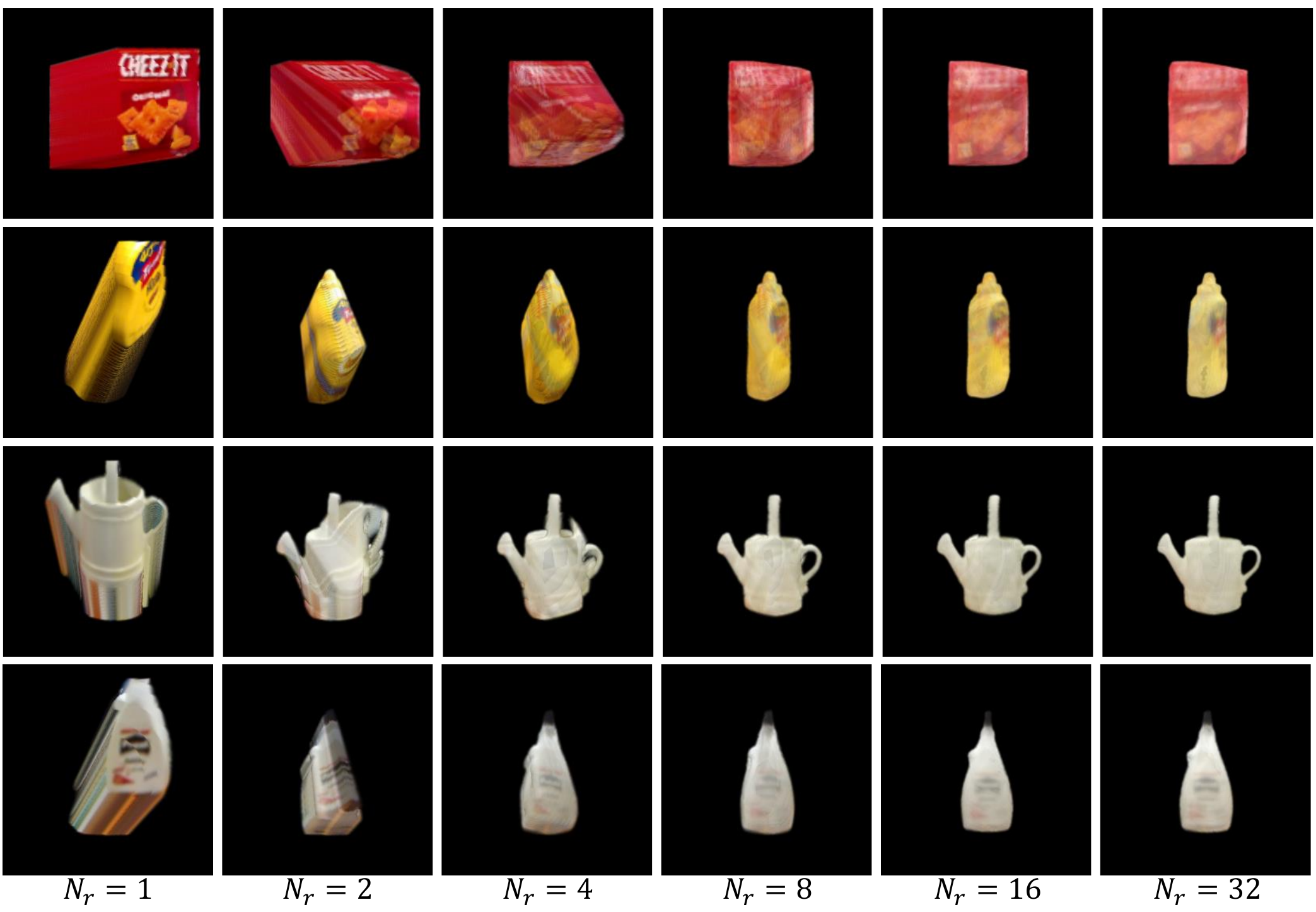}
\caption{\textbf{Qualitative results of $N_{r}$}: We visualize the projection of reference feature to compare the quality with the number of reference views. The first and second rows show the reference features from the YCBV dataset object, and the third and last rows visualize the reference feature from the LM dataset object.}
\label{fig:qualitative_ref}
\end{figure}

We demonstrate the additional example result of each iteration for the LMO and YCBV dataset in Fig. \ref{fig:LMO_iter} and Fig. \ref{fig:YCBV_iter}, respectively. In addition, we illustrate additional qualitative results of the LMO dataset and YCBV dataset in Fig. \ref{fig:LMO_example} and Fig. \ref{fig:YCBV_example}, respectively. Additionally, to demonstrate the competence of our method, we compare the qualitative results of our method and the other state-of-the-art methods in \ref{fig:qualitative_other}. We also visualize reference feature quality in \ref{fig:qualitative_ref} with the number of references. As shown in the figure, the more the reference view is used, the more accurate the shape becomes, but there is a trade-off in which the texture is blurred.

\newpage
\end{document}